\pdfoutput=1

\documentclass[11pt]{article}

\usepackage[final]{acl}

\usepackage{times}
\usepackage{latexsym}
\usepackage[normalem]{ulem}
\usepackage[T1]{fontenc}

\usepackage[utf8]{inputenc}

\usepackage{microtype}

\usepackage{inconsolata}

\usepackage{graphicx}

\usepackage{tabularx}
\usepackage{amsmath}
\usepackage{graphicx} 
\usepackage{enumitem}
\usepackage{booktabs}
\usepackage{diagbox}
\usepackage{float}
\usepackage{multirow}
\usepackage{booktabs}
\usepackage{algorithm}
\usepackage{algorithmic} 
\usepackage{colortbl} 
\usepackage{amssymb}
\definecolor{deepred}{rgb}{0.7, 0.0, 0.0}
\definecolor{lightblue}{rgb}{0.8, 0.9, 0.95}
\definecolor{deepblue}{rgb}{0.36, 0.44, 0.58}
\definecolor{deepgreen}{rgb}{0.38, 0.49, 0.38}
\title{CDS: Knowledge Component-Driven Data Synthesis Guided by Cognitive Diagnosis Theory}

\author{
    \begin{tabular}[t]{c}
        Haokun Zhao\textsuperscript{\rm 1},
        Jinyi Han\textsuperscript{\rm 2},
        Jiaqing Liang\textsuperscript{\rm 3 \thanks{Corresponding Author}} \\
        Yanghua Xiao\textsuperscript{\rm 1,2},
        Xiaojun Meng\textsuperscript{\rm 4},
        Jiansheng Wei\textsuperscript{\rm 4}
    \end{tabular} \\
    \textsuperscript{\rm 1}College of Computer Science and Artificial Intelligence, Fudan University\\
    \textsuperscript{\rm 2}Shanghai Institute of Artificial Intelligence for Education, East China Normal University\\
    \textsuperscript{\rm 3}School of Data Science, Fudan University \quad \textsuperscript{\rm 4}Huawei Noah's Ark Lab \\
    \texttt{hkzhao23@m.fudan.edu.cn, \{liangjiaqing,shawyh\}@fudan.edu.cn} \\
    \texttt{haixiahan03@gmail.com, \{xiaojun.meng,weijiansheng\}@huawei.com}
}

\begin{document}
\maketitle

\begin{abstract}
Large Language Models (LLMs) have achieved significant advancements, but the increasing complexity of tasks and higher performance demands highlight the need for continuous improvement. Some approaches utilize synthetic data generated by advanced LLMs based on evaluation results to train models. However, conventional evaluation methods fail to provide detailed, fine-grained profiles of LLMs, limiting their guidance for data synthesis. In this paper, we introduce the \textbf{\underline{C}ognitive \underline{D}iagnostic \underline{S}ynthesis} (CDS) method, which incorporates a diagnostic process inspired by \textbf{Cognitive Diagnosis Theory} (CDT) to refine evaluation results and characterize model profiles at the knowledge component level. Based on these diagnostics, we propose two diagnosis-synthesis strategies for weakness-targeted data synthesis. Additionally, we present an enhanced data augmentation and selection pipeline to improve the quality and diversity of synthesized data. Our experiments with several open-source models show significant improvements across multiple benchmarks, achieving up to 6.00\% improvement in code generation, 13.10\% in mathematical reasoning, and 5.43\% in academic exams. Code and data are available on GitHub \footnote{https://anonymous.4open.science/r/cds-04D1}.
\end{abstract}
\section{Introduction}
Large Language Models (LLMs) have demonstrated remarkable capabilities across diverse tasks. However, the increasing complexity of emerging tasks and the limitations revealed in real-world applications highlight the critical need for continuous improvement of LLM performance. 

To achieve continuous improvement, researchers typically analyze the model's evaluation metrics, then refine or supplement the corpora accordingly for subsequent iterations \cite{Lee2024LLM2LLMBL,Zhao2024CEMAD}. For example, if LLMs are found to perform poorly in mathematical tasks, more math data will be deliberately integrated into the dataset for the next training cycle. In this process, advanced LLMs (\textit{e.g.}, GPT-4) are increasingly utilized as data synthesizers to automate training data generation and augmentation \cite{Dai2023AugGPTLC,Liu2023LogiCoTLC,Sun2023PrincipleDrivenSO}, thereby reducing reliance on costly manual annotation.
\begin{figure}[t]
  \centering
  \includegraphics[width=0.48\textwidth]{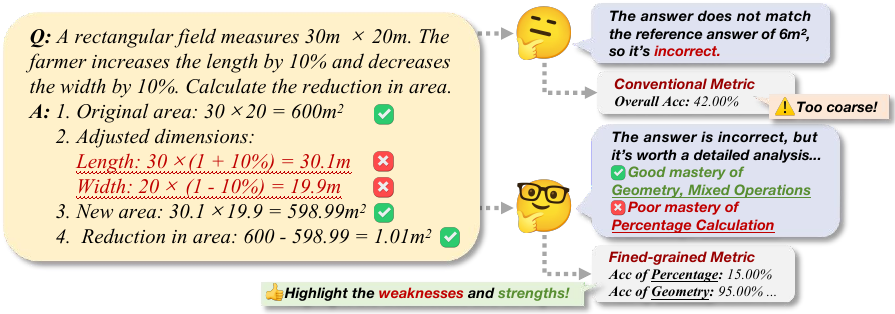}
  \caption{
The math problem assesses Geometry, Mixed Operations, and Percentage Calculations. The model performs well overall but makes an error in percentage calculation. Conventional metrics lack the granularity to capture these deficiencies, whereas fine-grained metrics can identify specific strengths and weaknesses at the knowledge level.}
  \label{fig:intro}
\end{figure}

However, there are two limitations in this process: 1) \textbf{Coarse-Grained Evaluation.} Conventional metrics, such as overall accuracy, focus solely on binary (correct/incorrect) outcomes for each test sample, providing a summary of model performance at the dataset level. Figure \ref{fig:intro} highlights the limitation of these metrics: when a model performs well on most of the knowledge assessed but makes a mistake in percentage calculation, leading to an incorrect final result, conventional metrics would simply classify such cases as “incorrect,” failing to identify the weakness in percentage calculation. As a result, the lack of granular evaluation limits the attribution of errors to specific sub-skills or competencies, which are referred to as \textbf{Knowledge Components} (KCs) in educational theory \cite{Moore2024AutomatedGA}, thus hindering the precise identification of weaknesses. 2) The \textbf{coarse-grained evaluation limits the guidance for subsequent data synthesis,} leading to generated data being general and insufficiently targeted at the specific weaknesses of the model. Recent studies attempt to use erroneous questions as seed data for synthesis to align the generated data with the model's weaknesses \cite{Lee2024LLM2LLMBL, Ying2024LLMsasInstructorsLF}, but they still treat each error in isolation, failing to map and summarize observed mistakes to underlying capability deficiencies. Thus, these methods may correct superficial errors but fail to address fundamental weaknesses at KC level.

To address these limitations, we draw inspiration from \textbf{Cognitive Diagnosis Theory} (CDT)—an educational framework that uses “diagnosis” to systematically map assessments to mastery of KCs, identifying specific strengths and weaknesses in students' abilities. In this paper, the LLMs to be enhanced are treated as “student.” We apply this diagnostic approach to summarize their performance in evaluations, profiling their capabilities at the KC level. This profiling then guides advanced LLMs to synthesize data aimed at improving the weak KCs in the next training cycle.



Specifically, we introduce the \textbf{\underline{C}ognitive \underline{D}iagnostic \underline{S}ynthesis} (CDS) method. First, we propose two diagnosis-synthesis strategies from different diagnostic perspectives, using advanced LLMs as data synthesizers: \textbf{1) Global Strategy:} Diagnosis at the dataset level with fine-grained metrics such as KC accuracy. These metrics quantify mastery of each KC, helping identify weak mastery KCs and generating tailored training data. \textbf{2) Fine-grained Strategy:} Diagnosis at the question level, leveraging the analytical capabilities of advanced LLMs \cite{Bai2023BenchmarkingFM,Dai2023AugGPTLC}. We use advanced LLMs to perform cognitive diagnosis on specific erroneous cases, identifying KCs requiring remediation. These analyses generated during the diagnostic process are then integrated into the synthesis prompt to expand the length of the chain-of-thought (CoT), as long CoTs enhance generation quality \cite{Jin2024TheIO,Wang2024DRTo1OD}.

These synthetic data will undergo augmentation through data rewriting and fusion to enhance their diversity and comprehensiveness. Following this, we propose a \textbf{two-stage data selection} process to ensure data quality. In Stage 1, an advanced LLM is used to filter out erroneous data. In Stage 2, a novelty score, \textbf{CDS\textsubscript{score}}, is designed, which references global diagnosis outcomes to select high-quality, weakness-relevant data.

Our contributions are as follows:
\begin{itemize}[noitemsep]
    \item We introduce the diagnosis process of CDT to refine conventional evaluation, using fine-grained knowledge components to characterize model capabilities.
    \item We propose two diagnosis-synthesis strategies from different diagnostic perspectives to achieve targeted data synthesis.
    \item We propose an improved data augmentation and selection pipeline to enhance the quality and diversity of synthesized data. Specifically, we introduce a novelty score, CDS\_score, enabling efficient selection of high-quality and relevant data.
    \item We conduct extensive experiments spanning multiple benchmarks and diverse domains, demonstrating the dominant effectiveness and applicability of CDS.
\end{itemize}

\section{Related Work}
\subsection{Cognitive Diagnosis Theory}
Cognitive Diagnosis Theory (CDT) provides fine-grained assessments by diagnosing an individual's mastery of specific knowledge points, offering actionable insights for targeted interventions \cite{Junker2001CognitiveAM,Rupp2010DiagnosticMT}. CDT focuses on identifying strengths and weaknesses through models such as DINA \cite{de2009dina} and G-DINA \cite{Torre2011TheGD}. These models leverage Q-Matrix Theory \cite{Tatsuoka1983RULESA} to link test items with underlying knowledge points and provide probabilistic mastery estimates. While CDT integrated with AI has been widely applied in educational assessments \cite{Minn2022AIassistedKA,Wang2019NeuralCD,liu2021towards}, its application in data synthesis and model improvement is highly underexplored.

\subsection{Synthetic Data for Improving Model}


Leveraging advanced LLMs to generate training data has become a widely adopted strategy for improving open-source models \cite{Dai2023AugGPTLC, Xu2023WizardLMEL, Mitra2024OrcaMathUT, Wang2023HowFC, Ivison2023CamelsIA, chen2023alpagasus, Mitra2023Orca2T, Fu2023SpecializingSL, Kumar2020DataAU, Li2024Common7L, Li2023TunaIT}. Concurrently, researchers have investigated generating corrective data through error analysis of target models \cite{An2023LearningFM, Lee2024LLM2LLMBL} and enhancing learning via comparative analysis of positive and negative examples \cite{Ying2024LLMsasInstructorsLF}. \citet{Zhang2024InContextPL} optimized prompts by extracting reasoning principles from errors, while \citet{Liao2024NeuralSymbolicCD} analyzed errors in smaller LMs, storing derived knowledge and summaries in specialized knowledge bases to enhance reasoning performance.

Some studies begin with knowledge-based synthesis, generating knowledge concepts from online course platforms \cite{Huang2024MUSTARDMU}, GPT-4 \cite{Li2024SyntheticD}, and seed instruction analysis and clustering \cite{Huang2024KeyPointDrivenDataSynthesisEnhancement}, thereby guiding advanced LLMs in data synthesis. However, these approaches have several limitations: simple nominal concepts are inadequate for producing high-quality and diverse synthetic data and may significantly deviate from real-world distributions. Moreover, these methods focus solely on synthesis and overlook the potential of knowledge points to evaluate model weaknesses, thereby limiting the targeting and effectiveness of data synthesis.

\section{CDS Method}
\label{sec:methods}

\begin{figure*}[htb]
  \centering
  \includegraphics[width=1.0\textwidth]{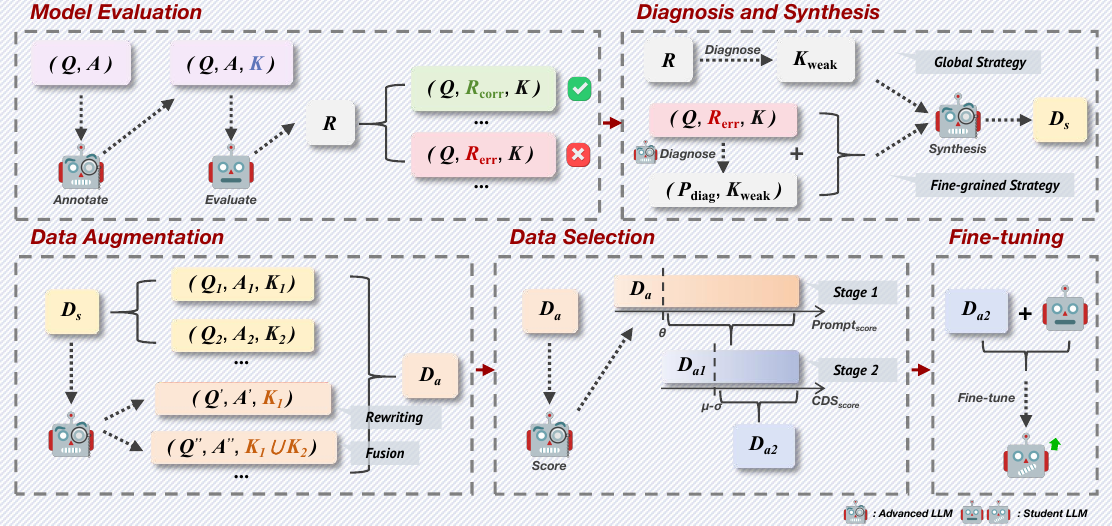}
  \caption{The pipeline of CDS method.}
  \label{fig:pipeline}
\end{figure*}

In the framework shown in Figure \ref{fig:pipeline}, we annotate the test samples in the benchmark with their KCs to evaluate the model and diagnose the evaluation results to identify weak KCs, which serve as targets to guide subsequent data synthesis. These synthetic data will undergo augmentation and selection processes to ensure quality, and then be used for supplementary training of the student model. The details are provided as follows.

\subsection{Model Evaluation}
\label{sec:eval}
\textbf{KC Annotation.} For a benchmark $\mathcal{D} = \{d \mid d = (q, a_\text{ref})\}$, where each sample $d$ consists of a question $q$ and a reference answer $a_\text{ref}$, we allocate the training data to the target dataset, $\mathcal{D}_\text{target}$, and reserve the test data for the evaluation dataset, $\mathcal{D}_\text{eval}$. An advanced model $\mathcal{M}_\text{a}$ is then used to annotate each sample in $\mathcal{D}_\text{target}$ with relevant KCs. The annotation process is carried out in two stages.

\textbf{Stage 1: } We use $\mathcal{M}\text{a}$ to perform \textbf{coarse annotations}, with the prompt shown in Figure \ref{fig:1}. To ensure an appropriate level of granularity, we sample chapter titles from digital learning platforms, such as MOOCs, to provide examples of KCs like Probability. Indeed, subsequent experiments will demonstrate the flexibility of KC annotations. We aggregate the KC tags from each sample to build an initial set. This set is then refined by $\mathcal{M}_\text{a}$, with optional expert involvement, to eliminate redundancies and ensure that the KCs are mutually exclusive, collectively exhaustive, and appropriately granular. The refined set of KCs is denoted as $\mathcal{K}$.

\textbf{Stage 2: } We use $\mathcal{M}_\text{a}$ to perform \textbf{constrained annotations} to ensure that the tagged KCs originate from $\mathcal{K}$, yielding the tagged benchmark $\mathcal{D}^{*}_\text{target}=\{d^{*}|d^{*}=(q,a_\text{ref},\mathcal{K}_\text{q}),\mathcal{K}_\text{q}\subsetneq \mathcal{K}\}$. Thereby, we build the Question-Knowledge Component ($Q\text{-}\mathit{KC}$) matrix, represented as:

\small
\noindent
\begin{flalign}
&Q\text{-}\mathit{KC} \in \{0, 1\}^{|\mathcal{D}_\text{target}| \times |\mathcal{K}|} \\
&Q\text{-}\mathit{KC} [i,j] =
\begin{cases}
1 & \text{if } kc_j \in \mathcal{K}_{\text{q}_i}, kc_j \in \mathcal{K}, \\
0 & \text{otherwise.}
\end{cases}
\end{flalign}
\noindent
where \( |\mathcal{D}_\text{target}| \) is the number of benchmark questions, \( |\mathcal{K}| \) is the cardinality of the KC set, and \( \mathcal{K}_{\text{q}_i} \) denotes the KC set tagged to question \( q_i \), which is a subset of \( \mathcal{K} \).
\normalsize 

\textbf{Model Evaluation and Result Collection.} We evaluate the student model $\mathcal{M}_\text{s}$ using the tagged benchmark $\mathcal{D}^{*}_\text{target}$. To better reveal the model's KC deficiencies, we collect \textbf{erroneous cases}, denoted as $\mathcal{D}_\text{err} = \{(q, r_\text{err}, \mathcal{K}_\text{q})\}$, for subsequent diagnosis, where $r_\text{err}$ is the response from $\mathcal{M}_\text{s}$ to question $q$ that does not match the reference answer $a_\text{ref}$.

\subsection{Diagnosis and Data Synthesis} 
Diagnosis is defined as the process of analyzing KC mastery profiles derived from evaluation results. To holistically assess the model's proficiency in KCs, we propose two diagnosis-synthesis strategies from different diagnostic perspectives.

\textbf{Global Strategy.}  
We diagnose the model's performance from a global perspective using aggregated KC metrics at the dataset level. Specifically, we follow the \textit{Deterministic Input, Noise And} (DINA) cognitive diagnosis framework \cite{de2009dina} that posits a binary mastery assumption:  
\begin{itemize}[noitemsep]
    \item A correct response to question $ q_i $ implies mastery of \textbf{all} associated KCs  
    \item An incorrect response implies \textbf{no mastery} of \textbf{any} associated KCs  
\end{itemize}

Under these assumptions, we compute KC-specific accuracy and frequency metrics:  
\small
\noindent
\begin{flalign}
&\text{Acc}(kc_j) = \frac{\sum_{i=1}^{|\mathcal{D}_\text{target}|} \mathbb{I}(q_i) \cdot Q\text{-}KC[i,j]}{\sum_{i=1}^{|\mathcal{D}_\text{target}|} Q\text{-}KC[i,j]} \label{eq:global_acc } \\
&\text{Freq}(kc_j) = \frac{\sum_{i=1}^{|\mathcal{D}_\text{target}|} Q\text{-}KC[i,j]}{|\mathcal{D}_\text{target}|} 
\end{flalign}
\noindent
where $\mathbb{I}(q_i) \in \{0,1\}$ indicates the correctness of $\mathcal{M}_\text{s}$'s response to $q_i$.
\normalsize

Based on these metrics, we construct a global KC diagnostic profile for $\mathcal{M}_\text{s}$, which consists of the accuracy and frequency of each KC, identifying \textbf{weakly mastered KCs} as those with low accuracy or low frequency, denoted as $\mathcal{K}_\text{w}$. To address these weaknesses, we use \(\mathcal{M}_\text{a}\) to generate data targeted at these weaknesses, based on \(\mathcal{K}_\text{w}\). The process is detailed in Algorithm~\ref{alg:global_strategy}.

\begin{algorithm}[h]
\small
\caption{Global Strategy}
\label{alg:global_strategy}
\begin{algorithmic}[1]
\REQUIRE A set of KCs $\mathcal{K}$, an advanced model $\mathcal{M}_\text{a}$, accuracy threshold $\delta_{\text{a}}$, frequency threshold $\delta_{\text{f}}$
\ENSURE A synthesized dataset $\mathcal{D}_{\text{global}}$

\STATE $\mathcal{K}_\text{w} \gets \emptyset$ 
\FOR{each $kc \in \mathcal{K}$}
    \IF{$\mathrm{Acc}(kc) \le \delta_{\text{a}}$ \textbf{or} $\mathrm{Freq}(kc) \le \delta_{\text{f}}$}
        \STATE $\mathcal{K}_\text{w} \gets \mathcal{K}_\text{w} \cup \{kc\}$
    \ENDIF
\ENDFOR

\STATE $\mathcal{D}_{\text{global}} \gets \emptyset$
\FOR{each $kc \in \mathcal{K}_\text{w}$}
    \STATE $(q,\,a,\,\mathcal{K}_q = \{kc\})
    \gets \mathrm{Generate}(\mathcal{M}_\text{a},\, kc)$
    \STATE $\mathcal{D}_{\text{global}} 
    \gets \mathcal{D}_{\text{global}} 
    \cup \{(q,\, a,\, \mathcal{K}_q)\}$
\ENDFOR

\RETURN $\mathcal{D}_{\text{global}}$
\end{algorithmic}
\normalsize
\end{algorithm}



This strategy operates exclusively at the KC level, avoiding the introduction of original questions into synthetic prompts. By summarizing specific questions into KCs, we address a key limitation of traditional example-question-based synthesis methods: when prompted with original questions, the model tends to unconsciously rewrite or rephrase them. This ensures that the generated data is both novel and independent, free from overfitting to the original dataset. Detailed prompts and case studies are provided in the Appendix. 

\textbf{Fine-grained Strategy.} We diagnose the model's \textbf{erroneous cases} at the question level from a fine-grained perspective. We use $\mathcal{M}_\text{a}$ as a diagnoser to analyze the model's problem-solving process in the current case, thereby identifying the underlying weak KCs exposed by this case, which are also denoted as $\mathcal{K}_\text{w}$. These analytical processes generated during the diagnosis are denoted as $p_\text{diag}$, and are integrated with the original questions and erroneous responses, forming long CoTs within the prompt for the advanced LLM to stimulate deeper reasoning, leading to higher-quality generation results. The process is detailed in Algorithm~\ref{alg:fine_grained}.


\begin{algorithm}[h]
\small
\caption{Fine-grained Strategy}
\label{alg:fine_grained}
\begin{algorithmic}[1]
\REQUIRE A set of erroneous responses $\mathcal{D}_\text{err} = \{\,d_\text{err}\mid d_\text{err}=(q, r_\text{err}, \mathcal{K}_q)\}$, an advanced model $\mathcal{M}_\text{a}$
\ENSURE A synthesized dataset $\mathcal{D}_{\text{fine-grained}}$

\STATE $\mathcal{O}_{\text{diag}} \gets \emptyset$ 
\COMMENT{\(\mathcal{O}_{\text{diag}}\) stores erroneous cases and their corresponding diagnostic outputs.}
\FOR{each $d_\text{err} \in \mathcal{D}_{\text{err}}$}
    \STATE $(p_{\text{diag}}, \mathcal{K}_{\text{w}}) 
    \gets \mathrm{Generate}\bigl(\mathcal{M}_\text{a},\; d_\text{err}\bigr)$
    \quad 
    \STATE $\mathcal{O}_{\text{diag}} 
    \gets \mathcal{O}_{\text{diag}} \cup 
    \bigl\{(d_\text{err},\,p_{\text{diag}},\,\mathcal{K}_{\text{w}})\bigr\}$
\ENDFOR

\STATE $\mathcal{D}_{\text{fine-grained}} \gets \emptyset$
\FOR{each $(d_\text{err},\,p_{\text{diag}},\,\mathcal{K}_{\text{w}}) \in \mathcal{O}_{\text{diag}}$}
    \STATE $(q',\, a',\, \mathcal{K}_{\text{w}})
    \gets \mathrm{Generate}\bigl(\mathcal{M}_\text{a},\; d_\text{err},\; p_{\text{diag}},\; \mathcal{K}_{\text{w}}\bigr)$
    \quad 
    \STATE $\mathcal{D}_{\text{fine-grained}}
    \gets \mathcal{D}_{\text{fine-grained}} \cup \{(q', a', \mathcal{K}_{\text{w}})\}$
\ENDFOR

\RETURN $\mathcal{D}_{\text{fine-grained}}$
\end{algorithmic}
\normalsize
\end{algorithm}



This diagnosis-synthesis paradigm leverages LLMs' analytical capabilities beyond data generation~\cite{Bai2023BenchmarkingFM,Dai2023AugGPTLC}. Recent studies show that long CoTs guide advanced LLMs to deeper reasoning, generating higher-quality outputs~\cite{Jin2024TheIO,Wang2024DRTo1OD}. By integrating diagnostic processes into data synthesis, our strategy produces more targeted and higher-quality data than direct synthesis methods. Detailed prompts and case studies are in the Appendix.

\textbf{Data Augmentation.} We concatenate the data generated by two synthesis methods and employ the following augmentation strategies to further increase data diversity and volume:
\begin{itemize}[noitemsep]
    \item \textbf{KC-Constrained Rewriting:} Adapting traditional data rewriting methods \cite{Dai2023AugGPTLC,Sun2023PrincipleDrivenSO}, we add a constraint: the rewritten data should contain the same KCs as the originals, avoiding deviation from targeted weaknesses while enhancing diversity.
    \item \textbf{Multi-KC Fusion:} We pair data from the synthetic dataset and prompt the advanced LLM to generate new data containing KCs from both, thereby increasing data complexity and comprehensiveness.
\end{itemize}

We sample a small proportion of the data for augmentation and reintegrate the augmented samples into the dataset. Additionally, we limit the max number of KCs per data to prevent the generation of overly complex or ambiguous samples.

\subsection{Data Selection}
We implement a two-stage data selection process to refine augmented synthetic dataset $\mathcal{D}_\text{a}$, eliminating subpar samples and retaining those that meet high-quality and high-relevance standards.

\textbf{Stage 1: }$\mathcal{M}_\text{a}$ assigns scores to the data based on multiple criteria such as correctness and KC relevance, filtering out samples below a threshold.

\textbf{Stage 2:} Simply scoring data individually with the model does not leverage global KC diagnostic profiles to select weakness-targeted data. Thus, we introduce a novel metric, CDS\textsubscript{score}. We hypothesize that data with more KCs have higher complexity and comprehensiveness, while data containing low-frequency and low-accuracy KCs are more effective for targeting weaknesses. Based on these assumptions, specifically, for \( d_\text{a} = (q, a, \mathcal{K}_q) \in \mathcal{D}_\text{a} \), the CDS\textsubscript{score} is calculated as follows:


\small 
\noindent
\begin{flalign}
&\mathcal{V}(kc_j) = w_1 \log (\text{Acc}(kc_j) + \epsilon) + w_2 \log (\text{Freq}_{a}(kc_j) + \epsilon) \\
&\mathit{CDS}_\text{score}(d_a) = \sum_{kc_j \in \mathcal{K}_q} \mathcal{V}(kc_j) 
\end{flalign}
\noindent
where \( \mathcal{V}(\cdot) \) represents the significance of KC, \( \text{Acc}(kc_j) \) denotes the student LLM's initial accuracy on \( kc_j \), \( \text{Freq}_s(kc_j) \) is the frequency of \( kc_j \) in \( \mathcal{D}_\text{a} \), \( w_1 \) and \( w_2 \) are balancing weights, and \( \epsilon \) is a small constant to avoid division by zero.
\normalsize 

We apply a $1\text{-}\sigma$ principle, retaining samples with $\text{CDS}_\text{score}(d_i) > \mu - \sigma$, to construct the final training set for fine-tuning the student LLM.

\section{Experimental Setup}
\subsection{Datasets and Models}
\label{sec:dataset}

\textbf{Datasets.} We evaluate three primary tasks: mathematical reasoning, coding, and academic examination. To validate the effectiveness of CDS, we select GSM8k \cite{Cobbe2021TrainingVT}, MBPP \cite{Austin2021ProgramSW}, and GAOKAO-Bench \cite{Zhang2023EvaluatingTP} for each respective task. As described in Section \ref{sec:eval}, these benchmarks are split into $\mathcal{D}_\text{target}$ and $\mathcal{D}_\text{eval}$. Due to the similar distribution of In-Domain(ID) training and test sets, we also incorporate an Out-of-Domain(OOD) dataset in $\mathcal{D}_\text{eval}$ to better assess the generalization of CDS. The OOD datasets include GSM8k-PLUS \cite{Li2024GSMPlusAC}, HumanEval \cite{Chen2021EvaluatingLL}, and GAOKAO-Bench-Updates\footnote{https://github.com/OpenLMLab/GAOKAO-Bench-Updates}.

For the academic examination benchmark's KC annotation, we use chapter titles from the GAOKAO syllabus\footnote{https://gaokao.neea.edu.cn/html1/category/1509/6212-1.htm} as KCs, offering an alternative method distinct from Section \ref{sec:eval}.

\textbf{Models.} We use Llama3-8B-Instruct\cite{llama3modelcard} and Qwen1.5-7B-Chat\cite{Bai2023QwenTR} as the student LLMs, with Qwen2-72B-Instruct\cite{Yang2024Qwen2TR} serving as the advanced LLM.
\subsection{Setups}
\textbf{Training Setup.} We train the models on 1 NVIDIA A800 GPU using ZeRO Stage 1~\citep{20-zero} and AdamW~\citep{14-adamw} as the optimizer, with LoRA~\citep{Hu2021LoRALA} (rank $r=8$). The batch size is 32, the maximum sequence length is 2,048, and training runs for 1 epoch.

\textbf{Inference Setup.} For code generation, mathematical reasoning, and academic exams, we use greedy decoding with a maximum output of 512 tokens. For data generation, we use a temperature of 0.5, top-$p$ of 0.8, and a maximum output of 4096 tokens. All inference is conducted in a 0-shot setting. See Appendix \ref{app:exp} for more details.

\setlength\tabcolsep{3pt} 
\begin{table*}[htp]
\small
  \centering
\begin{tabular}{>{\arraybackslash}p{1.8cm}>{\centering\arraybackslash}p{1.25cm}>{\centering\arraybackslash}p{1.25cm}>{\centering\arraybackslash}p{1.25cm}>{\centering\arraybackslash}p{1.25cm}>{\centering\arraybackslash}p{1.25cm}>{\centering\arraybackslash}p{1.25cm}>{\centering\arraybackslash}p{1.25cm}>{\centering\arraybackslash}p{1.25cm}}
\toprule 
\toprule 
\multirow{3}{*}{\textbf{Method}} & \multicolumn{2}{c}{\textbf{Coding}}                & \multicolumn{2}{c}{\textbf{Math}}   & \multicolumn{2}{c}{\textbf{Examination}} &\textbf{Avg}\\ 
 \cmidrule(r{2pt}){2-3} \cmidrule(lr{2pt}){4-5}\cmidrule(lr{2pt}){6-7} \cmidrule(l{2pt}){8-8}  
 & \textbf{MBPP} & \textbf{H-Eval}  & \textbf{GSM8k} & \textbf{GSMPlus}   & \textbf{GAOKAO} & \textbf{GAOKAO\textsubscript{U}} & \multirow{2}{*}{\textbf{--}}\\ 
 \cmidrule(lr{2pt}){2-2} \cmidrule(lr{2pt}){3-3}\cmidrule(lr{2pt}){4-4}\cmidrule(lr{2pt}){5-5}\cmidrule(lr{2pt}){6-6}\cmidrule(lr{2pt}){7-7}
 & \textbf{P@1}& \textbf{P@1} & \textbf{Acc}  & \textbf{Acc} & \textbf{Acc}  & \textbf{Acc}  \\ \midrule

\rowcolor[rgb]{ .949,  .953,  .961}\multicolumn{8}{c}{\textit{Qwen1.5-7B-Chat}}     
\\
Prompt(vanilla)   &32.00    & 40.24   & 54.00  &33.92  &60.60  &48.87        &\cellcolor{lightblue}44.94  \\ \midrule
IFT           & 30.80          & 39.02                  & 52.42             & 33.72    &61.40 &47.96     &\cellcolor{lightblue}44.22    \\ \midrule
LEC            & 32.40          & 39.63                  & 52.20             & 33.76    &\textcolor{deepred}{\textbf{64.80}} &\textbf{53.39}     &\cellcolor{lightblue}46.03   \\ \midrule
AugGPT              & \textbf{34.80}            & 40.85                 & 46.38             & 28.86    &63.40  &52.49    &\cellcolor{lightblue}44.46  \\ \midrule
LLM2LLM             & 34.40  & \textbf{43.90}& 53.82  & 33.76  &64.40 &\textbf{53.39} &\cellcolor{lightblue}\textbf{47.28}    \\ \midrule

MUSTARD            & 35.40         & 39.02                & \textbf{57.42}         & \textbf{35.96}   &62.40 &52.04 &\cellcolor{lightblue}47.04    \\ \midrule
\textbf{CDS}(our)           & \textcolor{deepred}{\textbf{38.00}}                & \textcolor{deepred}{\textbf{44.51}}                  & \textcolor{deepred}{\textbf{64.54}}            & \textcolor{deepred}{\textbf{43.86}}  &\textbf{64.60} &\textcolor{deepred}{\textbf{53.85}}  &\cellcolor{lightblue}\textcolor{deepred}{\textbf{51.56}}\\  \midrule
\rowcolor[rgb]{ .949,  .953,  .961}\multicolumn{8}{c}{\textit{Llama3-8B-Instruct}}   
\\                                                                                                                                                                    
Prompt(vanilla)                        & 40.80             & 54.88   & 62.02 &  42.50   &41.20 &33.03 &\cellcolor{lightblue}45.74    \\ \midrule
IFT                   & 41.60 & \textbf{55.49}                  & 61.64             & \textbf{42.78}    &41.00 &31.67  &\cellcolor{lightblue}45.70     \\ \midrule
LEC                & \textcolor{deepred}{\textbf{42.80}}          & 54.88                  & 55.62        & 42.48      &\textbf{41.60} &32.58 &\cellcolor{lightblue}44.99\\  \midrule
AugGPT                      & 40.00 &51.83                  & 47.32             & 36.00  &\textcolor{deepred}{\textbf{42.20}} &\textbf{35.28}     &\cellcolor{lightblue}42.10    \\ \midrule
LLM2LLM                     & \textcolor{deepred}{\textbf{42.80}} &      \textcolor{deepred}{\textbf{56.10}}             & 55.76             & 41.46   &\textbf{41.60} & 31.67   &\cellcolor{lightblue}44.90  \\ \midrule

MUSTARD                    & 41.40 & 54.88               & \textbf{62.16}            &   42.44  &41.20&33.03  &\cellcolor{lightblue}\textbf{45.85}   \\ \midrule
\textbf{CDS}(our)                 & \textcolor{deepred}{\textbf{42.80}}    & \textbf{55.49}       & \textcolor{deepred}{\textbf{73.14}}                  & \textcolor{deepred}{\textbf{55.60}}    &41.40 &\textcolor{deepred}{\textbf{38.46}}&\cellcolor{lightblue}\textcolor{deepred}{\textbf{51.15}}\\  \bottomrule\bottomrule 
\end{tabular}

 \caption{\label{tab:mainres} The main experimental results of our methods and baseline approaches across various tasks are presented. Experiments are conducted using two different LLMs: Qwen1.5-7B-Chat and Llama3-8B-Instruct. The top two performances are highlighted in \textcolor{deepred}{\textbf{red bold}} and \textbf{black bold}, respectively.}
 
\end{table*}


\subsection{Baselines}
\textbf{Main Experiments. } We consider several baselines for comparison with our method as follows:
\textit{(1) Prompt}: Direct prompting for answers. \textit{(2) IFT}: Fine-tuning with in-domain training data. \textit{(3) LEC} \cite{Ying2024LLMsasInstructorsLF}: Embedding erroneous cases with SentenceBERT \cite{Reimers2019SentenceBERTSE}, selecting similar positive examples via L2 distance, and synthesizing with both positive and negative cases. \textit{(4) AugGPT} \cite{Dai2023AugGPTLC}: Sampling unused instructions from the in-domain training set for synthetic data generation with the advanced LLM. \textit{(5) LLM2LLM} \cite{Lee2024LLM2LLMBL}: Generating additional data from incorrect examples using the advanced LLM. \textit{(6) MUSTARD} \cite{Huang2024MUSTARDMU}: Generating questions from seed concepts, followed by advanced LLM-generated answers and correctness filtering. Consistent data quantity is maintained across baselines: 2k for mathematics, 0.5k for code generation, and 0.5k for academic examination.

\textbf{Data Selection Experiments.} We evaluate our data selection algorithm against several baselines as follows:
\textit{(1) CBS} \cite{Chen2023MaybeO0}: Instructions are embedded using SentenceBERT, clustered with HDBSCAN \cite{Campello2013DensityBasedCB}, and selected using the K-Center-Greedy algorithm. 
\textit{(2) Coreset} \cite{Sener2017ActiveLF}: Similar to CBS, with instructions embedded using SentenceBERT and selected using K-Center-Greedy. 
\textit{(3) Diversity} \cite{Wang2022SelfInstructAL}: For each data, the ROUGE score is computed against a subset of $n$ samples, and the $k$ with the lowest ROUGE scores are selected. 
\textit{(4) Length}: Samples are selected based on data length, focusing on the longest instances (Length\textsubscript{long}).
\textit{(5) Perplexity} \cite{marion2023moreinvestigatingdatapruning}: Samples are selected based on low per-token perplexity, indicating model certainty and fluency.
\textit{(6) AlpaGasus} \cite{chen2024alpagasustrainingbetteralpaca}: Instances are scored by an advanced LLM like ChatGPT on dimensions such as helpfulness and accuracy, and low-scoring instances are filtered out. 
\textit{(7) Random}: Instances are randomly selected from the dataset.

\setlength\tabcolsep{3pt}
\begin{table*}[!htp]
\small
  \centering
    \begin{tabular}{>{\arraybackslash}p{1.8cm}cccccccccccccccc}
    \toprule \toprule 
    \multirow{3}{*}{\textbf{Method}} & \multicolumn{7}{c}{\textbf{Coding}} & \multicolumn{7}{c}{\textbf{Math}} \\ 
    \cmidrule(lr{2pt}){2-8} \cmidrule(lr{2pt}){9-15}
    & \multicolumn{3}{c}{\textbf{MBPP}} & \multicolumn{3}{c}{\textbf{H-Eval}} &\textbf{Avg}& \multicolumn{3}{c}{\textbf{GSM8k}} & \multicolumn{3}{c}{\textbf{GSMPlus}}&\textbf{Avg} \\ 
    \cmidrule(lr{2pt}){2-4} \cmidrule(lr{2pt}){5-7} \cmidrule(lr{2pt}){8-8}\cmidrule(lr{2pt}){9-11} \cmidrule(lr{2pt}){12-14}\cmidrule(lr{2pt}){15-15}
    & \textbf{0.1k} & \textbf{0.2k} & \textbf{0.3k} & \textbf{0.1k} & \textbf{0.2k} & \textbf{0.3k} & \textbf{--} &\textbf{0.4k} & \textbf{0.8k} & \textbf{1.6k} & \textbf{0.4k} & \textbf{0.8k} & \textbf{1.6k} & \textbf{--}\\ 
    \midrule
    CBS           &25.60 &33.80 &33.40 &\textcolor{deepred}{\textbf{43.29}} &\textcolor{deepred}{\textbf{38.41}} &37.20 &\cellcolor{lightblue}35.28   & 41.80          & \textbf{62.00}          & 60.94 & 27.08                  & \textcolor{deepred}{\textbf{40.40}}             & 40.24    &\cellcolor{lightblue}45.41          \\ \midrule
    CoreSet     &25.60 &33.40 &35.60 &40.85 &34.15 &39.63 &\cellcolor{lightblue}34.87   & 50.76          & 59.70          & 60.06 & 34.10                  & 38.14             & 38.66         &\cellcolor{lightblue}46.90      \\ \midrule
    Diversity     &26.60 &34.40 &35.40 &\textbf{42.68} &32.32 &\textbf{42.07} &\cellcolor{lightblue}35.58    & 43.28          & 56.14          & 61.64 & 28.54                  & 35.08             & 39.80       &\cellcolor{lightblue}44.08      \\ \midrule
    Length\textsubscript{long} &23.80 &33.20 &33.20 &42.07 &34.15 &40.85   &\cellcolor{lightblue}34.55          & 36.96          & 59.52          & 61.02 & 24.10                  & 38.40             & 39.46       &\cellcolor{lightblue}43.24        \\ \midrule
    Perplexity   &\textcolor{deepred}{\textbf{31.80}} &\textcolor{deepred}{\textbf{35.00}} &35.00 &\textbf{42.68} &\textbf{37.20} &41.46  &\cellcolor{lightblue}\textbf{37.19}        & \textbf{53.04} & 60.80          & \textbf{62.52} & \textbf{35.32} & 40.18 & \textcolor{deepred}{\textbf{41.56}}  &\cellcolor{lightblue}\textbf{48.90} \\ \midrule
    AlpaGasus    &30.60 &34.20 &35.40 &40.24 &35.37 &40.85  &\cellcolor{lightblue}36.11            & 49.44          & \textcolor{deepred}{\textbf{62.34}} & 56.82 & 30.70 & \textbf{40.20} & 37.32     &\cellcolor{lightblue}46.14     \\ \midrule
    Random        &24.60 &34.00 &35.20 &40.24 &34.15 &39.63  &\cellcolor{lightblue}34.64           & 45.68          & 58.38          & 59.32 & 33.76 & 38.70 & 38.86   &\cellcolor{lightblue}45.78   \\ \midrule

    \textbf{CDS\textsubscript{score}}(our) &\textbf{31.60} &\textbf{34.60} &\textcolor{deepred}{\textbf{36.20}} &40.85 &37.20 &\textcolor{deepred}{\textbf{43.29}}&\cellcolor{lightblue}\textcolor{deepred}{\textbf{37.29}} & \textcolor{deepred}{\textbf{53.80}} & 60.80 & \textcolor{deepred}{\textbf{62.56}} & \textcolor{deepred}{\textbf{36.02}} & 39.96 & \textbf{41.24}   &\cellcolor{lightblue}\textcolor{deepred}{\textbf{49.06}}   \\ 
    \bottomrule\bottomrule 
    \end{tabular}
  
 \caption{\label{tab:filterres} The experimental results of our data selection strategy and baseline approaches across various tasks are presented. Experiments are conducted using Qwen1.5-7B-Chat. The top two performances are highlighted in \textcolor{deepred}{\textbf{red bold}} and \textbf{black bold}, respectively.}
 
\end{table*}
\section{Experiments}
\subsection{Main Results}

The main experimental results of our methods
and baseline approaches across various tasks are pre-
sented in Table \ref{tab:mainres}. Our observations are summarized as follows:

\textbf{Dominant effectiveness and applicability of CDS.} CDS demonstrates significant improvements across different models and tasks. For example, on the GSM8k task, Qwen1.5-7B improves by 10.54\%, and Llama3-8B by 11.12\%. Additionally, CDS consistently produces optimal results in coding and examination tasks(including subjects such as biology, chemistry, geography, history, mathematics, and physics), highlighting its effectiveness and broad applicability to tasks that can be decomposed into well-defined KCs.

\textbf{Strong generalization.} Although CDS utilizes synthetic data generated based on ID tasks, it generalizes effectively to OOD tasks. For example, Qwen1.5-7B shows a 4.27\% improvement on the Humaneval Bench, while Llama3-8B improves by 5.43\% on GAOKAO-Bench-Updates. In contrast, methods like AugGPT and LEC fail to consistently improve performance on OOD tasks and may even lead to degradation.

\textbf{Flexibility in KC Annotation.} Section \ref{sec:dataset} explains that GAOKAO-Bench constructs the KC set for annotation by summarizing chapter titles, while the MBPP and GSM8k benchmarks use the annotation approach described in Section \ref{sec:eval}. Despite these different annotation methods, all benchmarks show consistent performance improvements with CDS.

\textbf{CDS's data selection improves robustness.} Unfiltered use of synthetic data may lead to model degradation. For instance, both LEC and AugGPT experienced performance declines on two mathematical benchmarks, with AugGPT showing a 7.62\% decrease on GSM8k and 5.06\% on GSMPlus. This aligns with prior research, which suggests that unchecked, low-quality instructional data can impair model performance \cite{Zhou2023LIMALI, chen2023alpagasus}. Table \ref{tab:errorcase} shows an erroneous sample generated by AugGPT, which can be filtered out during CDS's selection Stage 1.

\subsection{Evaluation of Selection Strategies}
\setlength\tabcolsep{3pt} 
\begin{table*}[!htp]
\small
  \centering
\begin{tabular}{>{\arraybackslash}p{3.05cm}>{\centering\arraybackslash}p{1.25cm}>{\centering\arraybackslash}p{1.25cm}>{\centering\arraybackslash}p{1.25cm}>{\centering\arraybackslash}p{1.25cm}>{\centering\arraybackslash}p{1.25cm}>{\centering\arraybackslash}p{1.25cm}>{\centering\arraybackslash}p{1.25cm}>{\centering\arraybackslash}p{1.25cm}}
\toprule 
\toprule 
\multirow{2}{*}{\textbf{Strategy}} & \multicolumn{2}{c}{\textbf{Coding}}                & \multicolumn{2}{c}{\textbf{Math}}   & \multicolumn{2}{c}{\textbf{Examination}} &\textbf{Avg}\\ 
 \cmidrule(r{2pt}){2-3} \cmidrule(lr{2pt}){4-5}\cmidrule(lr{2pt}){6-7} \cmidrule(l{2pt}){8-8}  
 & \textbf{MBPP} & \textbf{H-Eval}  & \textbf{GSM8k} & \textbf{GSMPlus}   & \textbf{GAOKAO} & \textbf{GAOKAO\textsubscript{U}} & \multirow{2}{*}{\textbf{--}}\\ 
 \cmidrule(lr{2pt}){2-2} \cmidrule(lr{2pt}){3-3}\cmidrule(lr{2pt}){4-4}\cmidrule(lr{2pt}){5-5}\cmidrule(lr{2pt}){6-6}\cmidrule(lr{2pt}){7-7}
 & \textbf{P@1}& \textbf{P@1} & \textbf{Acc}  & \textbf{Acc} & \textbf{Acc}  & \textbf{Acc}  \\ \midrule
Prompt(vanilla) &32.00 &40.24   & 54.00  &33.92  &60.60  &48.87        &\cellcolor{lightblue}44.94  \\ \midrule
\rowcolor[rgb]{ .949,  .953,  .961}\multicolumn{8}{c}{\textit{Synthesis Strategy}}     
\\

+ Global         &32.00 & \textcolor{deepred}{\textbf{40.85}} &57.96 &36.22 & \textcolor{deepred}{\textbf{62.00}} &\textbf{52.04} &\cellcolor{lightblue}46.85   \\ \midrule
+ Fine-grained  &\textbf{34.00} &39.63 &\textbf{60.96} &\textbf{37.62} &60.80 &49.77     &\cellcolor{lightblue}\textbf{47.13}   \\ \midrule
+ Global \& Fine-grained & \textcolor{deepred}{\textbf{35.40}} &\textbf{40.24} & \textcolor{deepred}{\textbf{61.84}} & \textcolor{deepred}{\textbf{38.60}} &\textbf{61.40} & \textcolor{deepred}{\textbf{52.49}}  &\cellcolor{lightblue}\textcolor{deepred}{\textbf{48.33}}  \\ \midrule
\rowcolor[rgb]{ .949,  .953,  .961}\multicolumn{8}{c}{\textit{Augmentation Strategy}}     
\\
+ Rewrite &\textcolor{deepred}{\textbf{35.40}} &41.46 &\textbf{61.52} &\textbf{41.38} &62.60 &51.13  &\cellcolor{lightblue}48.91     \\ \midrule
+ Fusion &35.20 &41.46 &61.00 &40.42 &\textbf{63.40} &\textbf{52.49} &\cellcolor{lightblue}\textbf{48.99}\\  \midrule
+ Rewrite \& Fusion &\textcolor{deepred}{\textbf{35.40}} &\textcolor{deepred}{\textbf{43.29}} &\textcolor{deepred}{\textbf{63.78}} &\textcolor{deepred}{\textbf{42.04}} &\textcolor{deepred}{\textbf{64.60}} &\textcolor{deepred}{\textbf{53.85}}  &\cellcolor{lightblue}\textcolor{deepred}{\textbf{50.49}}    \\ \bottomrule\bottomrule 
\end{tabular}

\caption{\label{tab:strategy} Ablation results of data synthesis and augmentation strategies. Experiments were conducted using Qwen1.5-7B-Chat. Data augmentation was applied to the synthetic data generated by the `+ Global \& Fine-grained' approach. The top two performances are highlighted in \textcolor{deepred}{\textbf{red bold}} and \textbf{black bold}, respectively. The sample sizes for the synthesis strategy are 300, 300, and 600 for coding; 300, 1000, and 1300 for math; and 300, 300, and 600 for exams. The augmentation proportions are 0.5, 0.5, 0.25 \& 0.25.}
 
\end{table*}
Experimental results comparing different data selection strategies are presented in Table \ref{tab:filterres}, where we fine-tuned Qwen1.5-7B-Chat using samples selected from an augmented synthetic dataset (pre-screened by Qwen2-72B-Instruct for basic correctness). To evaluate the effectiveness of the CDS score, we retained only Stage 2 data selection and excluded Stage 1, which is similar to AlpaGasus. Our key findings are:

\textbf{High-quality data selection and broad task applicability.} CDS\textsubscript{score} achieves the best average metrics on both math and coding tasks. Specifically, in 12 tests corresponding to three different sample sizes across four datasets, 8 tests ranked in the top 2. CDS\textsubscript{score} consistently outperforms the Random method in all scenarios, with average improvements of 2.65\% and 3.28\% for the two tasks, respectively. This demonstrates that CDS\textsubscript{score} enhances training data quality through selection and exhibits broad task applicability.

\textbf{Strong stability.} CDS\textsubscript{score} shows consistent performance across datasets and sample sizes, while some baselines exhibit fluctuating performance. For example, CBS's performance varied significantly with sample size, and Coreset performed well on math tasks but struggled with coding tasks.

\textbf{Scalability.} CDS\textsubscript{score} expands its advantage as sample sizes increase, demonstrating optimal accuracy with 0.3k samples for MBPP and H-Eval, and 1.6k samples for GSM8K. Notably, at 0.3k samples on Humaneval, it outperformed the second-best methods, Diversity and Perplexity, by 1.22\% and 1.83\%, respectively.

\textbf{Computational Efficiency.} As shown in Table \ref{tab:timecost}, compared to the suboptimal Perplexity method when selecting 2,000 samples from a 3,000-sample dataset, Perplexity requires 307.78 seconds, while CDS\textsubscript{score} takes negligible time. This efficiency results from the elimination of computationally intensive tasks such as embedding generation and clustering, and without the need for GPU.

\subsection{Ablation Study}
To evaluate the effectiveness of each component of the CDS method, we conducted an ablation study using various combinations of data synthesis and augmentation strategies. A basic correctness check was performed using Qwen-72B-Instruct to filter out obviously incorrect data. The results are shown in Table \ref{tab:strategy}. Our key findings are:

\textbf{Dual Strategies Outperform Single Strategies.} Dual strategies generally outperform single strategies in both the data synthesis and augmentation stages. This effect is particularly noticeable in augmentation, where combining Rewriting and Fusion strategies resulted in optimal performance across all six datasets.

\textbf{Data Augmentation Effect.} Combining synthesis and augmentation typically improves performance. After applying augmentation, the average performance exceeded previous results. However, exceptions exist, such as with Rewriting, where performance on GAOKAO-UPDATES deteriorated by 1.36\%. This may be due to overfitting to similar synthesis data. Such degradation was not observed with dual strategy augmentation, suggesting that combining multiple strategies to increase data complexity and diversity improves robustness.

\section{Conclusion}


In this paper, we introduce the \textbf{\underline{C}ognitive \underline{D}iagnostic \underline{S}ynthesis} (CDS) method, inspired by Cognitive Diagnosis Theory, which refines evaluation results and characterizes model profiles at the knowledge component level. Building on the Knowledge Component diagnostic construct, we optimize conventional evaluation metrics, design dual diagnosis-synthesis strategies, and propose a novel data selection method. Leveraging advanced LLMs, we automate the generation of weakness-targeted, high-quality instructional data. We applied the synthesized data to small LLMs, such as Qwen1.5-7B-Chat and Llama3-8B-Instruct, achieving significant improvements in code generation, mathematical reasoning, and academic testing. Notably, CDS achieves these improvements without relying on expensive closed-source LLMs like GPT-4, instead using only the open-source Qwen2-72B-Chat for automated diagnosis and synthesis.

\section{Limitations}
In this paper, (1) due to cost limitations, our advanced LLM selection is restricted to the open-source Qwen-72B-Instruct, which is not the most cutting-edge model available. Given the current limitations in both the model's analytical and generative capabilities, the full potential of CDS remains to be explored. In future work, we plan to experiment with more advanced models, such as GPT-4\cite{Achiam2023GPT4TR} and DeepSeek-V3\cite{deepseekai2024deepseekv3technicalreport}.

(2) The identification and annotation of KCs still involve significant randomness and subjectivity. In future work, we aim to refine our KC annotation strategy, including exploring the use of pseudo-labels as a substitute for explicit labels.

\bibliography{custom}
\appendix
\section{Knowledge Component Details}

The knowledge components (KCs) used for various tasks are shown in Table~\ref{tab:kc_table_tasks} and Table~\ref{tab:kc_table_exams}. Table~\ref{tab:kc_table_tasks} lists the components for Coding and Math, while Table~\ref{tab:kc_table_exams} details the components for the various subjects within the Exams task.
\setlength\tabcolsep{8pt}
\begin{table*}[htp]
\small
  \centering
    \begin{tabular}{>{\arraybackslash}p{2cm} p{12cm}}
    \toprule \toprule 
    \textbf{Task} & \textbf{Knowledge Components} \\ 
    \midrule
    \textbf{Coding} & Basic Data Types, Bitwise Operations, Boolean Logic, Class Definitions, Comparison Operators, Conditional Statements, Copying and Deep Copying, Dictionary Operations, Dynamic Programming, Exception Handling, Finding Min and Max, Heap Operations, Importing Libraries and Modules, Indexing and Slicing, Lambda Functions, List and Array Operations, Looping, Map Function, Recursion, Regular Expressions, Search Algorithms, Sorting Algorithms, Stacks and Queues, String Operations, Summation, Tree Structures, Tuple Operations, Type Checking and Conversion \\ 
    \midrule
    \textbf{Math} & Basic Arithmetic Operations, Decimal and Fraction Operations, Mixed Operations, Prime and Composite Numbers, Factors and Multiples, GCD and LCM, Algebraic Expressions, Equations, Inequalities, Basic Geometry, Area, Perimeter, Volume, Angles, Coordinates, Mean, Median, Mode, Probability, Permutations, Combinations, Financial Calculations, Unit Conversion, Time and Date Calculations, Speed, Distance, and Time, Measurement, Money, Ratio and Proportion, Bar Graphs, Line Graphs, Number Sequences, Word Problems, Linear Equations, Simple Algebra, Pattern Recognition, Mathematical Logic, Shapes and Spatial Understanding, Symmetry, Congruence, Units of Measurement, Temperature, Length, Mass, Capacity \\ 
    \bottomrule \bottomrule 
    \end{tabular}
 \caption{Knowledge Components for Coding and Math tasks. The components for Exam tasks are in Table~\ref{tab:kc_table_exams}.}
\label{tab:kc_table_tasks}
\end{table*}

\begin{table*}[htp]
\small
  \centering
    \begin{tabular}{>{\arraybackslash}p{2.5cm} p{11.5cm}}
    \toprule \toprule 
    \textbf{Subject} & \textbf{Knowledge Components for Exams} \\ 
    \midrule
    \textbf{Biology} & Protein and Nucleic Acid Structure, Sugar and Lipid Types and Functions, Water and Inorganic Salts, Cell Theory, Prokaryotic and Eukaryotic Cells, Cell Membrane Structure, Organelles Structure and Function, Nucleus Structure and Function, Substance Transport Across Cell Membrane, Enzyme Role in Metabolism, ATP Metabolism, Photosynthesis Process, Environmental Impact on Photosynthesis, Cellular Respiration, Cell Growth and Division, Cell Differentiation, Cell Aging and Apoptosis, Cancer Cells and Prevention, Meiotic Division, DNA Structure and Replication, Gene Transcription and Translation, Mendel's Laws, Sex-linked Inheritance, Gene Mutation, Transgenic Food Safety, Human Genetic Diseases, Evolution Theory, Plant Hormones, Nervous and Hormonal Regulation, Nerve Impulse Transmission, Homeostasis, Immune System Role, Population and Community, Ecosystem Structure and Function, Ecosystem Stability, Biodiversity Conservation, Plant Growth Regulators, Yeast Respiration. \\
    \midrule
    \textbf{Chemistry} & Physical vs Chemical Changes, Acids, Bases, Salts and Oxides, Element Symbols, Valency and Formulas, Atomic and Molecular Masses, Law of Mass Conservation, Chemical Reactions, Molar Mass and Volume Calculations, Solubility and Concentration, Colloids and Solutions, Periodic Table Structure, Element Trends, Chemical Bonds, Oxidation-Reduction Reactions, Heat of Reactions, Electrochemistry, Reaction Rate and Activation Energy, Chemical Equilibrium, Electrolytes and Conductivity, pH Calculation, Ionization and Hydrolysis, Organic Compounds, Polymer Chemistry, Laboratory Safety, Gas Production and Separation, Chemical Analysis, Concentration Calculations. \\
    \midrule
    \textbf{Geography} & Earth's Position in Space, Solar Influence, Earth's Movements and Seasons, Earth's Layers, Earth Material Cycles, Surface Changes, Atmospheric Heating, Wind and Pressure Systems, Climate and Weather Systems, Water Cycle, Ocean Currents, Geography and Environment, Climate Change, Natural Resources, Natural Disasters, Population Growth, Migration, Urbanization, Agricultural and Industrial Location, Environmental Impact, Geography of Resources, Transportation Systems, Human-Earth Relationships, Sustainability, Green Development, Remote Sensing, Geographic Information Systems, GPS and Navigation, Digital Earth. \\
    \midrule
    \textbf{History} & Ancient Chinese Political Systems, Shang and Zhou Dynasties, Qin Centralization, Han to Yuan Political Evolution, Ming and Qing Monarchy, Ancient Chinese Economy, Agricultural Systems, Handicraft and Commerce, Capitalism Emergence, Cultural Evolution, Hundred Schools of Thought, Confucianism, Neo-Confucianism, Chinese Scientific and Technological Achievements, Ancient Greek and Roman Political Systems, Athenian Democracy, Roman Law, Renaissance, Enlightenment, Industrial Revolution, World War Effects, Cold War and Bipolarity, Globalization, WTO and China's Role, Modern Chinese Politics, Reform and Opening-up, Scientific and Technological Development in China, Modern Chinese Education and Culture. \\
    \midrule
    \textbf{Math} & Basic Arithmetic Operations, Decimal and Fraction Operations, Prime and Composite Numbers, Factors and Multiples, GCD and LCM, Algebraic Expressions, Equations, Inequalities, Geometry, Area, Perimeter, Volume, Angles, Coordinates, Mean, Median, Mode, Probability, Permutations, Combinations, Financial Calculations, Unit Conversion, Time and Date Calculations, Speed, Distance, Time, Measurement, Fractions, Decimals, Ratio, Proportion, Bar Graphs, Line Graphs, Probability Theory, Number Sequences, Word Problems, Time Tables, Integer Operations, Linear Equations, Simple Algebra, Pattern Recognition, Mathematical Logic, Shapes and Spatial Understanding, Symmetry, Congruence, Binomial Theorem, Conic Sections. \\
    \midrule
    \textbf{Physics} & Motion of Particles, Newton's Laws, Mechanical Energy, Projectile and Circular Motion, Law of Gravitation, Electric Field, Circuits, Magnetic Field, Electromagnetic Induction, AC, Kinetic Theory of Gases, States of Matter, Thermodynamics, Vibration and Waves, Electromagnetic Waves, Light, Relativity, Reference Frames, Linear Motion, Friction, Hooke’s Law, Scalars and Vectors, Force Composition, Equilibrium, Energy Conservation, Momentum, Collision Theory, Universal Gravitation, Satellite Motion, Electric Potential, Capacitors, Ohm’s Law, Electric Resistance, Magnetic Forces, Lorentz Force, Spectroscopy, Nuclear Physics, Radiation, Energy Loss, Fission and Fusion Reactions, Photoelectric Effect, Einstein's Equation. \\ 
    \bottomrule \bottomrule 
    \end{tabular}
 \caption{Knowledge Components for various Exam subjects. This table details the components for the Exam task mentioned in Table~\ref{tab:kc_table_tasks}.}
\label{tab:kc_table_exams}
\end{table*}

\section{Case Study}
\label{appendix:case}
\begin{table*}[htp]
\small
\centering
\begin{tabularx}{1.0\textwidth} { 
   >{\raggedright\arraybackslash}X 
   }
\toprule 
\toprule 
\textbf{An Erroneous Sample Generated by AugGPT}\\ 
  \midrule

\textbf{\#\#\# Question:} \\
A local bakery sells cupcakes at \$2 each. Last week, they decided to run a promotion where if you buy 5 cupcakes, you get 1 free. Without the promotion, they could sell 65 cupcakes in a week. After the promotion ended, the bakery noticed that their revenue increased by 40\% compared to the previous week when there was no promotion. How many free cupcakes did they give away during the promotion week?

\textbf{\#\#\# Answer:} \\
1. Calculate the number of cupcakes sold during the promotion week: \\Without the promotion, the bakery sold 65 cupcakes per week, so during the promotion week they sold: 65 * 1.4 = 91 cupcakes.\\2. Calculate the number of free cupcakes: Since the promotion is 'buy 5, get 1 free,' if X free cupcakes were given away during the promotion week, \textcolor{deepred}{\uwave{then: 5X + X = 91, so X = 15.33.}}\\Therefore, they gave away 15 free cupcakes.\\
 \midrule
 \textbf{\#\#\# Stage 1 Scoring Results:} \\
Score: 1 || Explanation:\\
	1.	Correctness of the Answer (Top Priority 1): The answer is incorrect. The model mistakenly assumes that the 91 cupcakes are all paid, but in reality, 91 represents the total number of paid cupcakes. Since the promotion is “buy 5, get 1 free,” for every 6 cupcakes sold, 1 is free. Therefore, the correct number of free cupcakes given away is 18, not 15.33, as 91 paid cupcakes correspond to 18 free cupcakes.\\
	2.	Clarity: The reasoning is presented in a step-by-step format but lacks accuracy in the core calculations. It is easy to follow but ultimately misleading because of the errors.\\
	3.	Conciseness: The explanation is brief but fails to address the problem correctly.\\
	4.	Format and Structure: The response is properly formatted in a clear, step-by-step structure, but the calculations are flawed.\\

Correctness was the most significant factor here, and the error in the calculations leads to a low score. The rest of the explanation could be clearer with correct reasoning.\\
\bottomrule\bottomrule
\end{tabularx}
\caption{\label{tab:errorcase} An erroneous sample generated by AugGPT. We used CDS's Stage 1 selection method, which employs Advanced LLM prompt scoring. The scoring results are shown above.}
\end{table*}
\section{Computational Cost of Data Selection Strategies}

As shown in Table \ref{tab:timecost}, we compared the time efficiency of different data selection methods when selecting 2,000 samples from a pool of 3,000. The key parameters and computational requirements are as follows:

\begin{itemize}
    \item \textbf{Model}: For embedding generation, we used the Qwen1.5-7B-Chat.
    \item \textbf{GPU Requirements}: 
        \begin{itemize}
            \item AlpaGasus requires 4 A800 GPUs and uses Qwen2-72B-Chat for inference.
            \item Other methods, including CBS, CoreSet, Diversity, and Perplexity, only require 1 A800 GPU for processing.
        \end{itemize}
    \item \textbf{Clustering Parameters}: 
        \begin{itemize}
            \item For CBS and CoreSet, the HDBSCAN clustering algorithm was used with the following parameters:
                \begin{itemize}
                    \item min\_cluster\_size = 2
                    \item min\_samples = 1
                \end{itemize}
        \end{itemize}
\end{itemize}

\setlength\tabcolsep{8pt}  
\begin{table*}[htp]
\small
  \centering
    \begin{tabular}{>{\arraybackslash}c c c c c c c c c}
    \toprule \toprule 
    \textbf{Method} & \textbf{CBS} & \textbf{CoreSet} & \textbf{Diversity} & \textbf{Length\textsubscript{long}} & \textbf{Perplexity} & \textbf{AlpaGasus} & \textbf{Random} & \textbf{CDS\textsubscript{score}}(our) \\ 
    \midrule
    Time (s) & 321.79 & 306.45 & 7134.14 & 0.00 & 307.78 & 1303.42 & 0.00 & 0.00 \\
    \bottomrule \bottomrule 
    \end{tabular}
  
 \caption{\label{tab:timecost} Time comparison of different data selection methods when selecting 2,000 samples from a pool of 3,000. A value of 0.00 indicates millisecond-level response time.}
\end{table*}

\section{Experimental Details}
\label{app:exp}
\subsection{Training Setup. }
We provide the key parameter settings for LoRA fine-tuning configurations. The following table \ref{tab:ftpara} summarizes the most important settings.

\begin{table}[htp]
\small
  \centering
    \begin{tabular}{>{\arraybackslash}c c}
    \toprule \toprule 
    \textbf{Parameter} & \textbf{Value} \\ 
    \midrule
    \textbf{Precision (bf16)} & Enabled \\ 
    \textbf{Optimizer} & AdamW \\ 
    \textbf{Learning Rate (lr)} & 3e-5 \\ 
    \textbf{Betas} & [0.98, 0.999] \\ 
    \textbf{Scheduler Type} & WarmupLR \\ 
    \textbf{Warmup Min LR} & 1e-4 \\ 
    \textbf{Warmup Max LR} & 3e-4 \\ 
    \textbf{Gradient Accumulation Steps} & 16 \\ 
    \textbf{Batch Size (per GPU)} & 2 \\ 
    \textbf{LoRA Rank (r)} & 8 \\ 
    \textbf{LoRA Alpha} & 16 \\ 
    \bottomrule \bottomrule 
    \end{tabular}
  
 \caption{Key parameter settings for the LoRA fine-tuning configuration.}
\label{tab:ftpara}
\end{table}

\subsection{Inference Setup}
We provide the key parameter settings for inference configurations. The following table \ref{tab:inference_setup} summarizes the most important settings.

\begin{table*}[htp]
\small
  \centering
    \begin{tabular}{l l}
    \toprule \toprule 
    \textbf{Stage} & \textbf{Sampling Parameters} \\ 
    \midrule
    \textbf{Dataset Annotation} & temperature=0.5, top\_p=0.8, repetition\_penalty=1.05, max\_tokens=1024 \\ 
    \textbf{Model Evaluation} & temperature=0, top\_p=1.0, top\_k=1, max\_tokens=512 \\ 
    \textbf{Fine-grained Diagnosis} & temperature=0.5, top\_p=0.8, repetition\_penalty=1.05, max\_tokens=1024 \\ 
    \textbf{Data Synthesis} & temperature=0.5, top\_p=0.8, repetition\_penalty=1.05, max\_tokens=4096, N\_sample=5 \\ 
    \textbf{Data Augmentation} & 
        \begin{tabular}[t]{@{}l@{}}
            temperature=0.5, top\_p=0.8, repetition\_penalty=1.05, max\_tokens=4096, \\
            p\_rw=0.25, p\_fusion=0.25
        \end{tabular} \\ 
    \textbf{Data Selection (Stage 1)} & 
        \begin{tabular}[t]{@{}l@{}}
            temperature=0, top\_p=1.0, top\_k=1, max\_tokens=512, \\
            repetition\_penalty=1.05, $\theta$=8
        \end{tabular} \\ 
    \textbf{Data Selection (Stage 2)} & w1=0.85, w2=0.15, $\epsilon$=1e-6 \\ 
    \bottomrule \bottomrule 
    \end{tabular}
  
 \caption{Key parameter settings for the inference configuration.}
\label{tab:inference_setup}
\end{table*}

\subsection{Dataset and Data Generation Details}
The dataset usage for the CDS method is summarized in Table~\ref{tab:dataset_usage}. It outlines the sample sizes used for each task, as well as the number of generated samples across various stages.
\begin{table*}[htp]
\small
  \centering
    \begin{tabular}{>{\arraybackslash}c c c c c c}
    \toprule \toprule 
    \textbf{Task} & \textbf{$\mathcal{D}_\text{target}$} & \textbf{$\mathcal{D}_\text{eval}$} & \textbf{$\mathcal{D}_\text{s}$} & \textbf{$\mathcal{D}_\text{a}$} & \textbf{$\mathcal{D}_\text{final}$} \\ 
    \midrule
    ID for Math:GSM8K* & 3500 & 5000 & 2581 & 3116 &2010 \\ 
    OOD for Math:GSMPLUS & - & 5000 & - & - & -\\ 
    \midrule
    ID for Coding:MBPP & 474 & 500 & 689 & 997 & 798\\ 
    OOD for Coding:H-Eval & - & 164 & - & - \\ 
    \midrule
    ID for Exams:GAOKAO & 591 & 500 & 1078 & 1776& 1332 \\ 
    OOD for Exams:GAOKAO\textsubscript{U} & - & 209 & - & -&- \\ 
    \bottomrule \bottomrule 
    \end{tabular}
  
 \caption{Dataset usage across different tasks and stages for Qwen1.5-7B-Chat. The "*" in the "ID for Math" task indicates that we additionally used 292 samples for pre-fine-tuning to help the model answer math problems in the required format, facilitating answer extraction.}
\label{tab:dataset_usage}
\end{table*}

\section{Instruction Details}
\begin{figure*}[htb]
  \centering
  \includegraphics[width=1.0\textwidth]{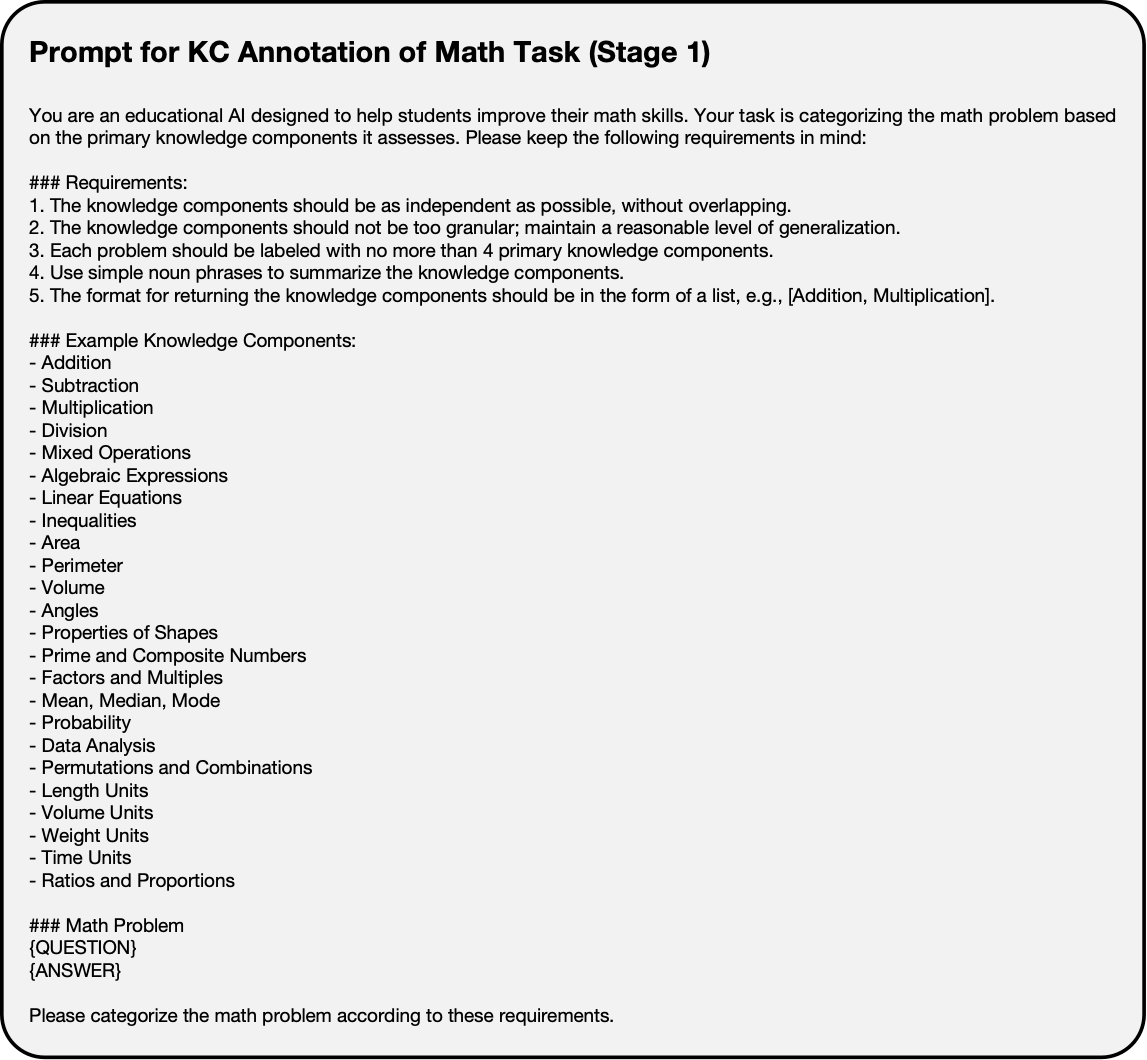}
  \caption{Prompt for KC Annotation of Math Task (Stage 1).}
  \label{fig:1}
\end{figure*}

\begin{figure*}[htb]
  \centering
  \includegraphics[width=1.0\textwidth]{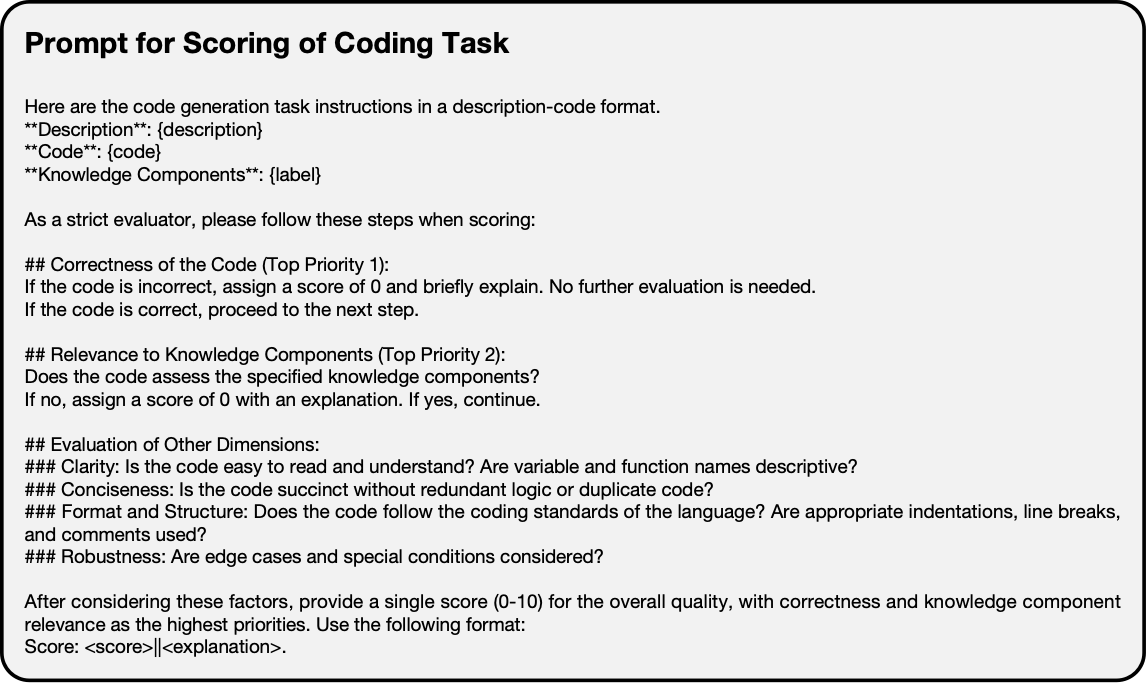}
  \caption{Prompt for Scoring of Coding Task.}
  \label{fig:16}
\end{figure*}

Table \ref{fig:1}~\ref{fig:16} show the prompts using in CDS.

\begin{figure*}[htb]
  \centering
  \includegraphics[width=1.0\textwidth]{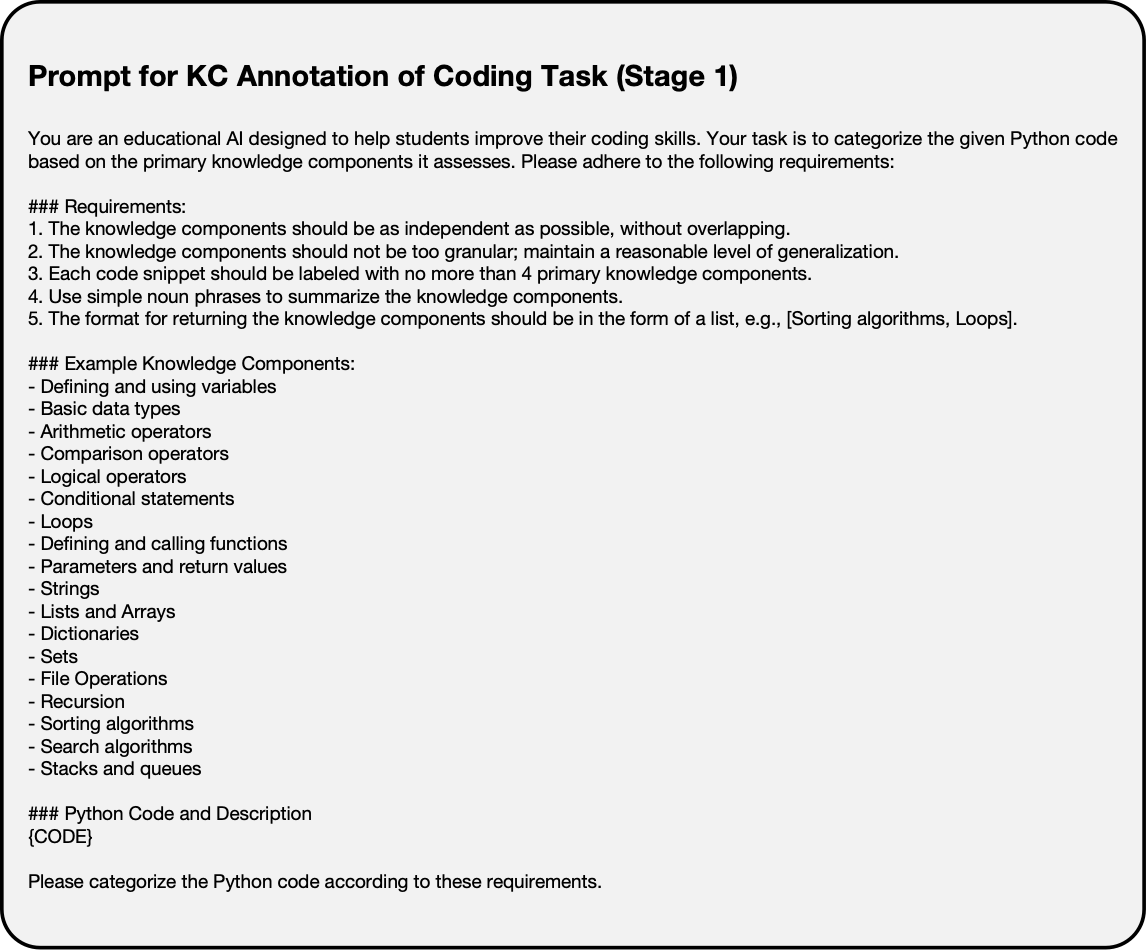}
  \caption{Prompt for KC Annotation of Coding Task (Stage 1).}
  \label{fig:2}
\end{figure*}

\begin{figure*}[htb]
  \centering
  \includegraphics[width=1.0\textwidth]{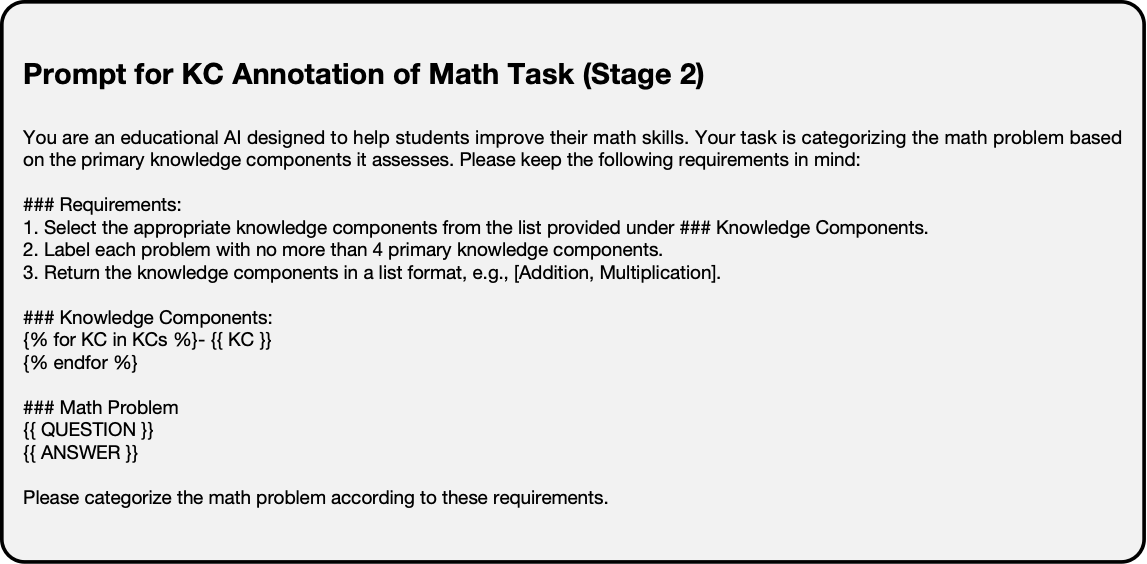}
  \caption{Prompt for KC Annotation of Math Task (Stage 2).}
  \label{fig:3}
\end{figure*}

\begin{figure*}[htb]
  \centering
  \includegraphics[width=1.0\textwidth]{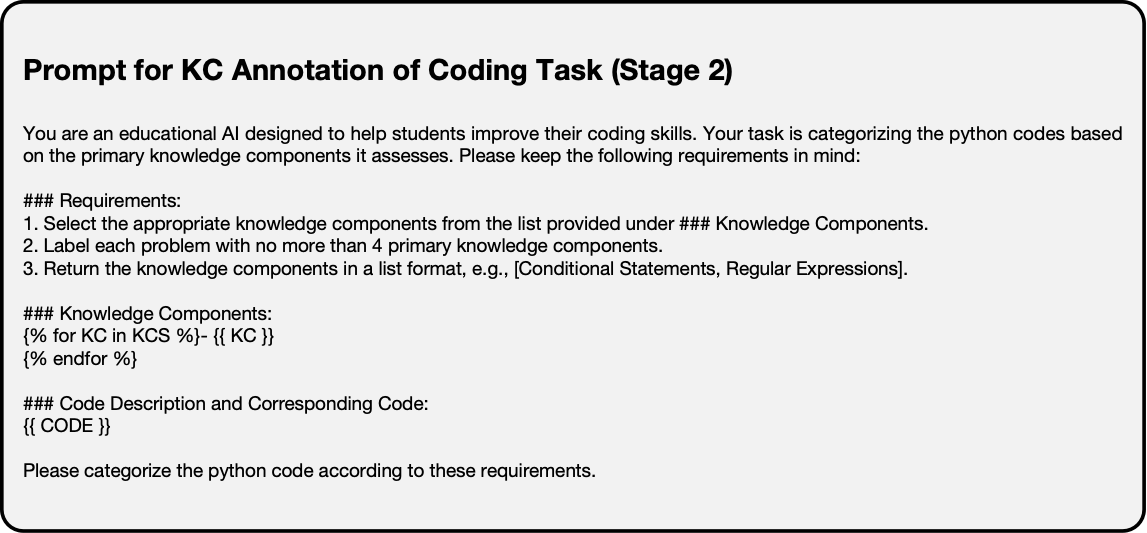}
  \caption{Prompt for KC Annotation of Coding Task (Stage 2).}
  \label{fig:4}
\end{figure*}

\begin{figure*}[htb]
  \centering
  \includegraphics[width=1.0\textwidth]{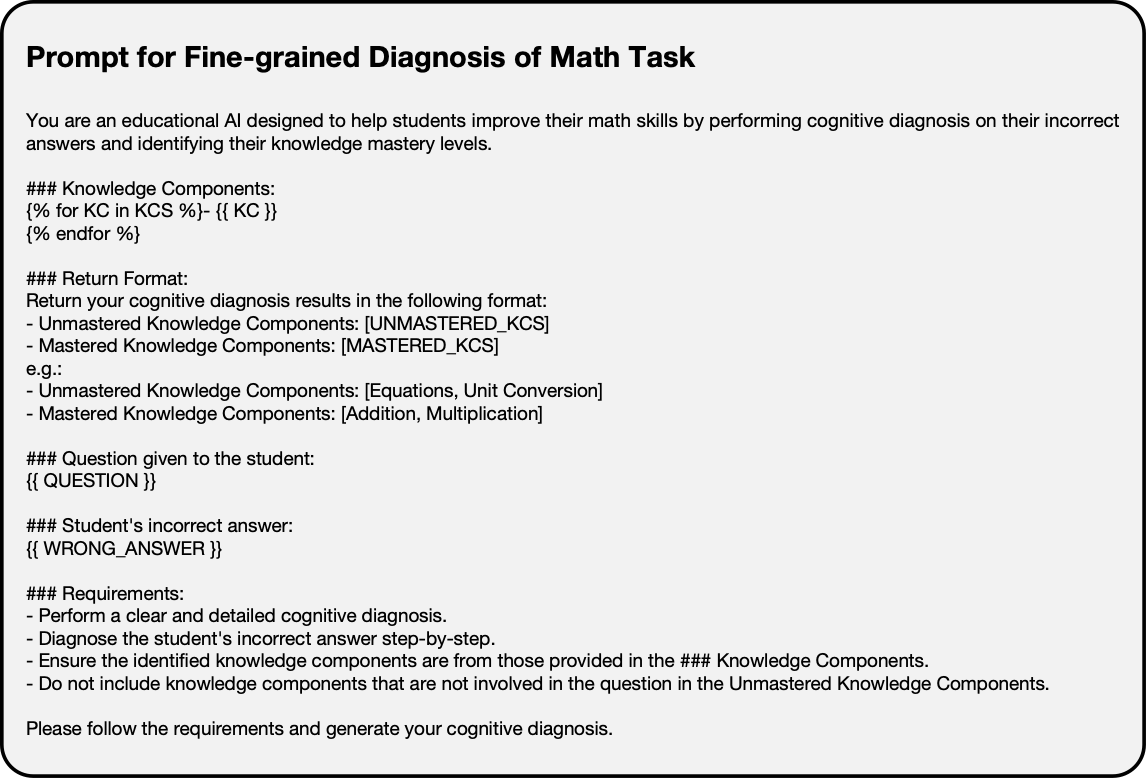}
  \caption{Prompt for Fine-grained Diagnosis of Math Task.}
  \label{fig:5}
\end{figure*}

\begin{figure*}[htb]
  \centering
  \includegraphics[width=1.0\textwidth]{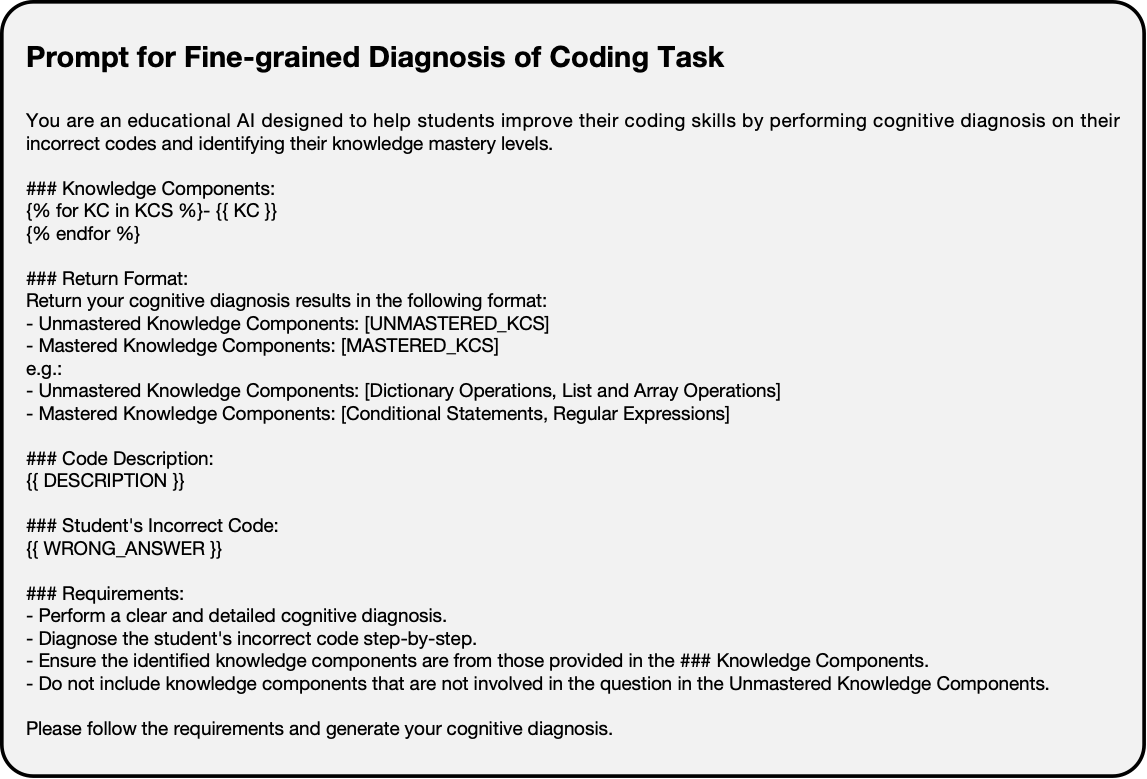}
  \caption{Prompt for Fine-grained Diagnosis of Coding Task.}
  \label{fig:6}
\end{figure*}

\begin{figure*}[htb]
  \centering
  \includegraphics[width=1.0\textwidth]{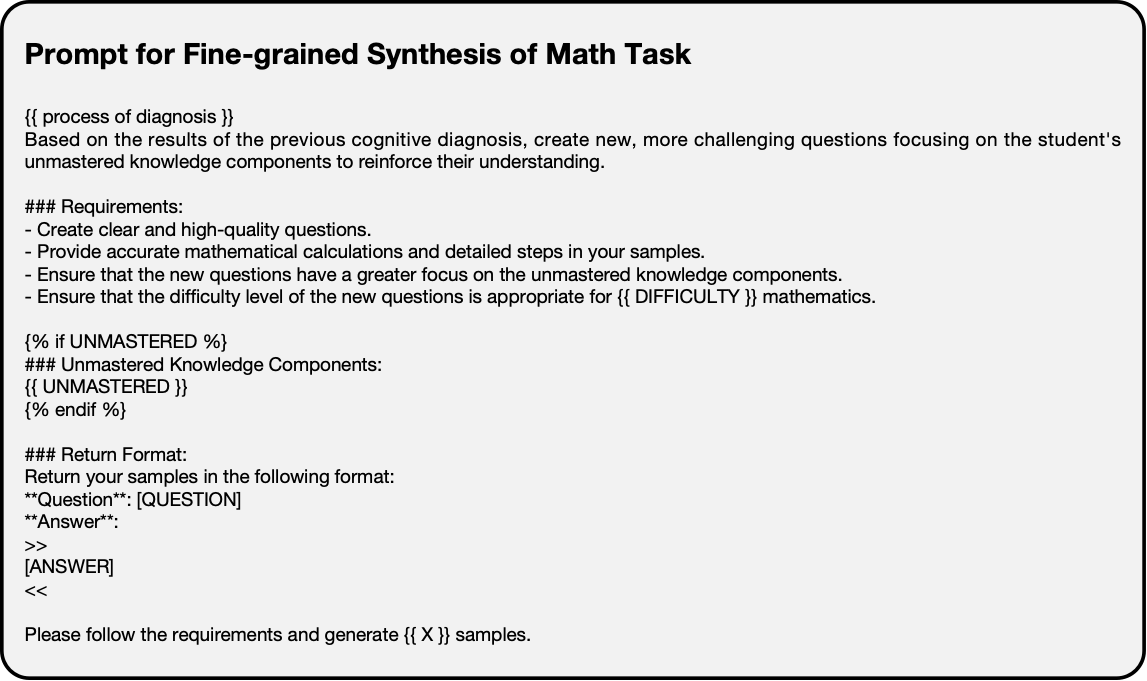}
  \caption{Prompt for Fine-grained Synthesis of Math Task.}
  \label{fig:7}
\end{figure*}

\begin{figure*}[htb]
  \centering
  \includegraphics[width=1.0\textwidth]{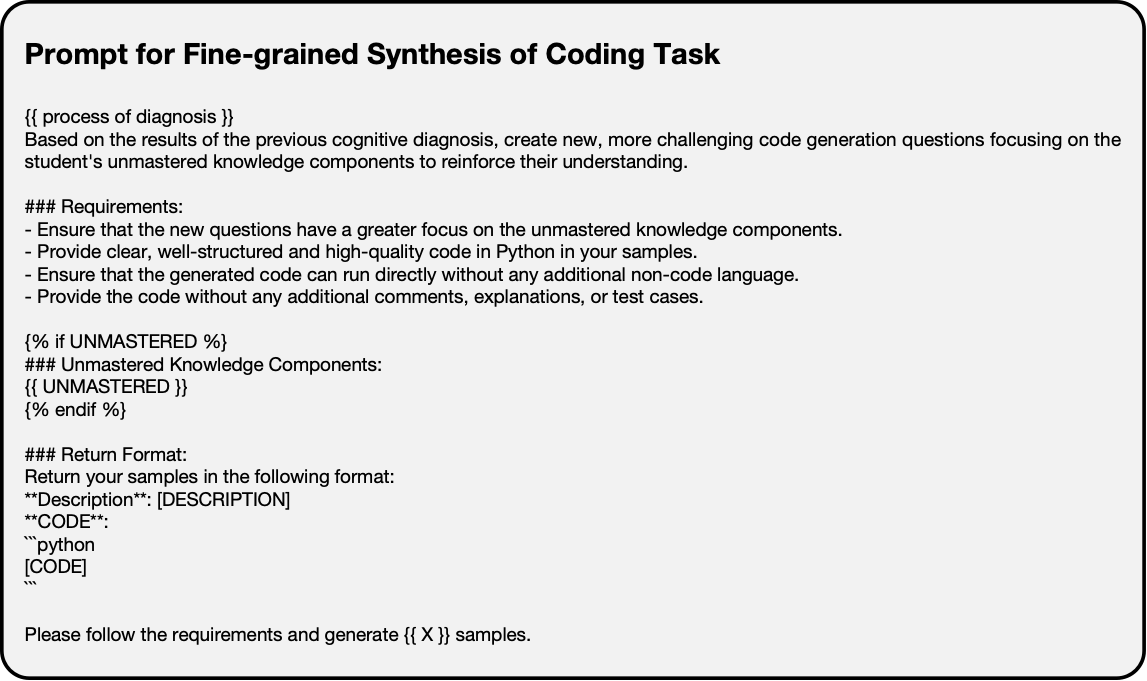}
  \caption{Prompt for Fine-grained Synthesis of Coding Task.}
  \label{fig:8}
\end{figure*}

\begin{figure*}[htb]
  \centering
  \includegraphics[width=1.0\textwidth]{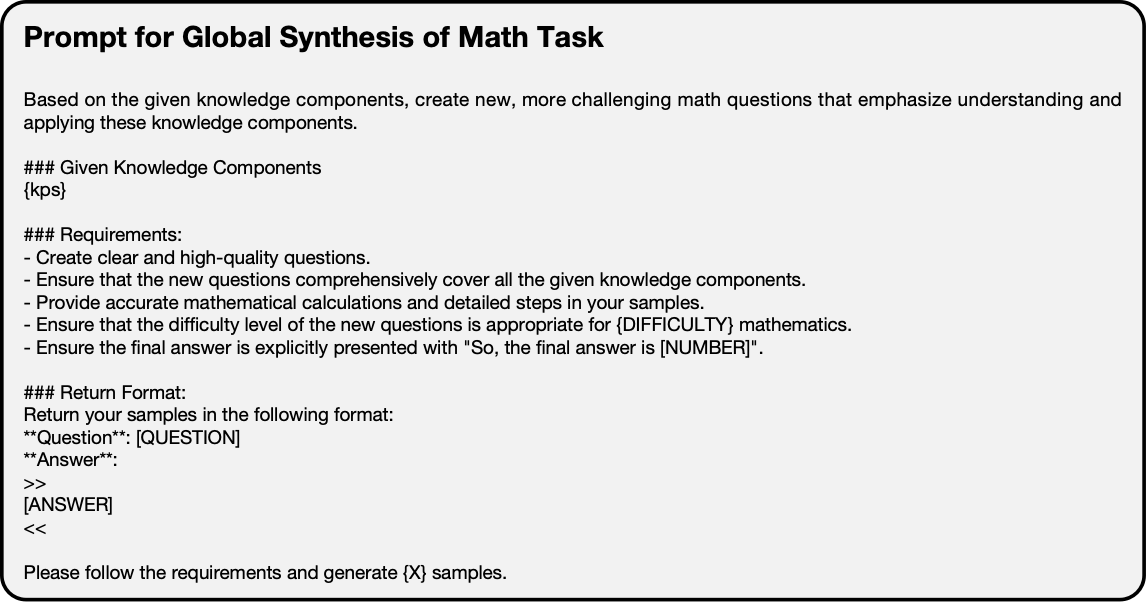}
  \caption{Prompt for Global Synthesis of Math Task.}
  \label{fig:9}
\end{figure*}

\begin{figure*}[htb]
  \centering
  \includegraphics[width=1.0\textwidth]{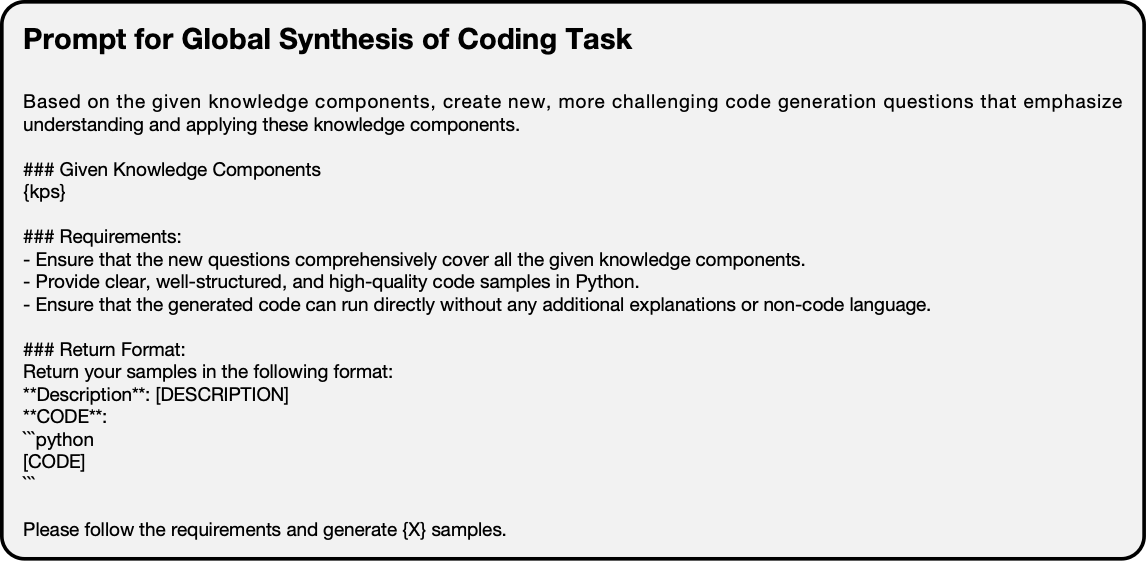}
  \caption{Prompt for Global Synthesis of Coding Task.}
  \label{fig:10}
\end{figure*}

\begin{figure*}[htb]
  \centering
  \includegraphics[width=1.0\textwidth]{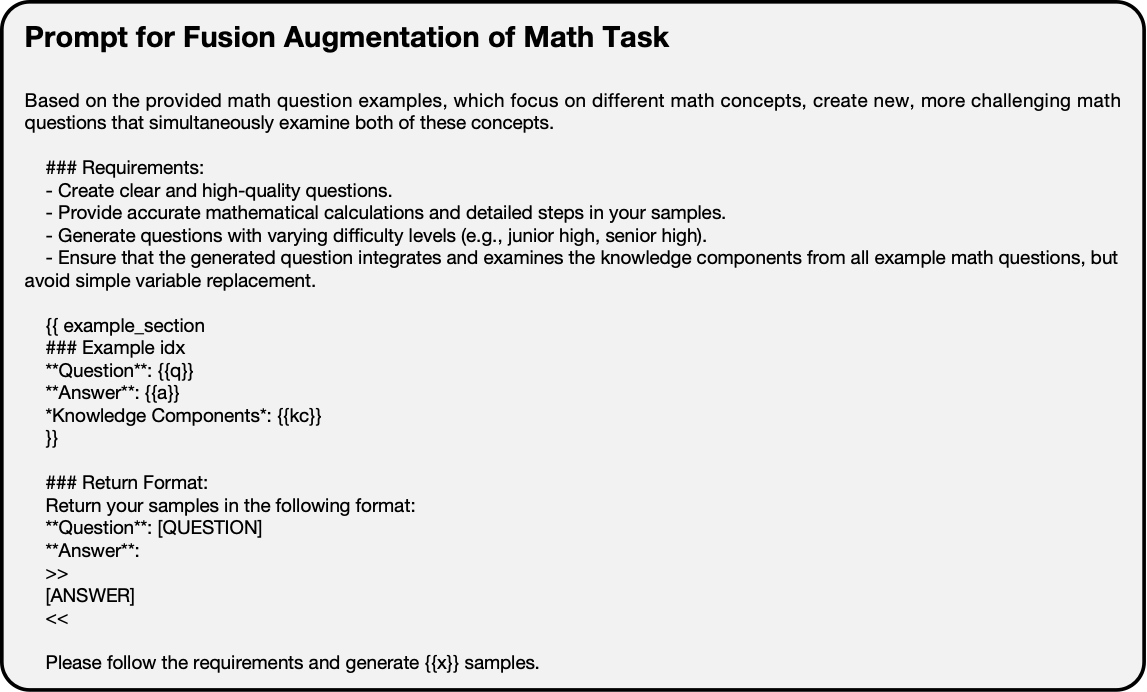}
  \caption{Prompt for Fusion Augmentation of Math Task.}
  \label{fig:11}
\end{figure*}

\begin{figure*}[htb]
  \centering
  \includegraphics[width=1.0\textwidth]{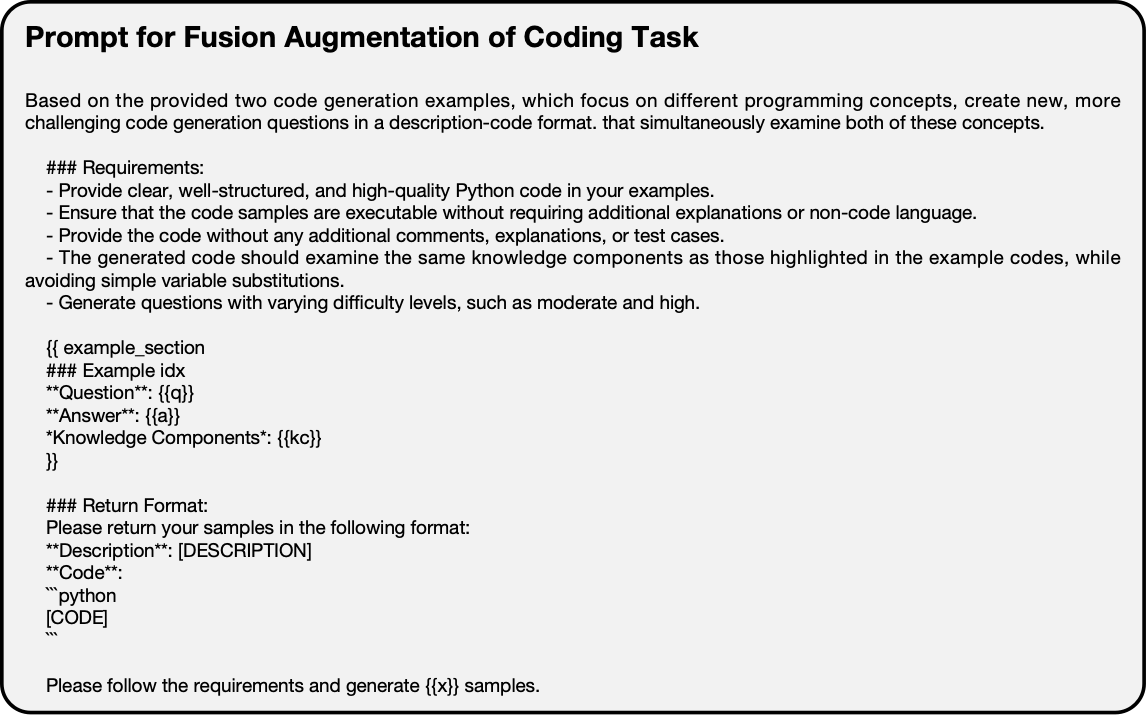}
  \caption{Prompt for Fusion Augmentation of Coding Task.}
  \label{fig:12}
\end{figure*}

\begin{figure*}[htb]
  \centering
  \includegraphics[width=1.0\textwidth]{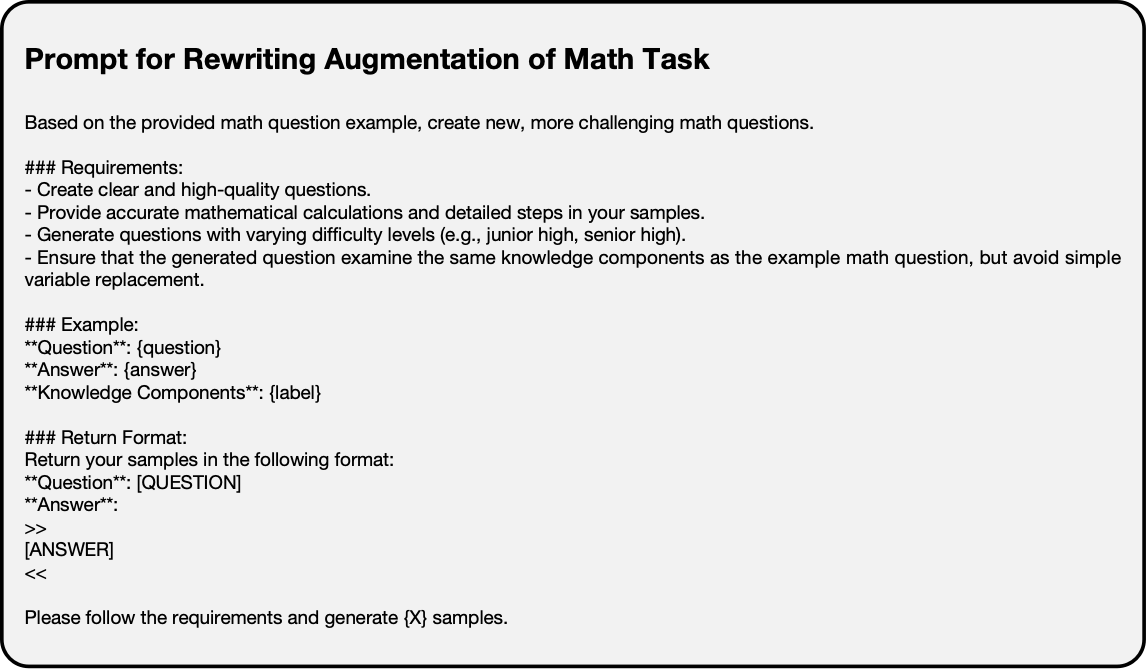}
  \caption{Prompt for Rewriting Augmentation of Math Task.}
  \label{fig:13}
\end{figure*}

\begin{figure*}[htb]
  \centering
  \includegraphics[width=1.0\textwidth]{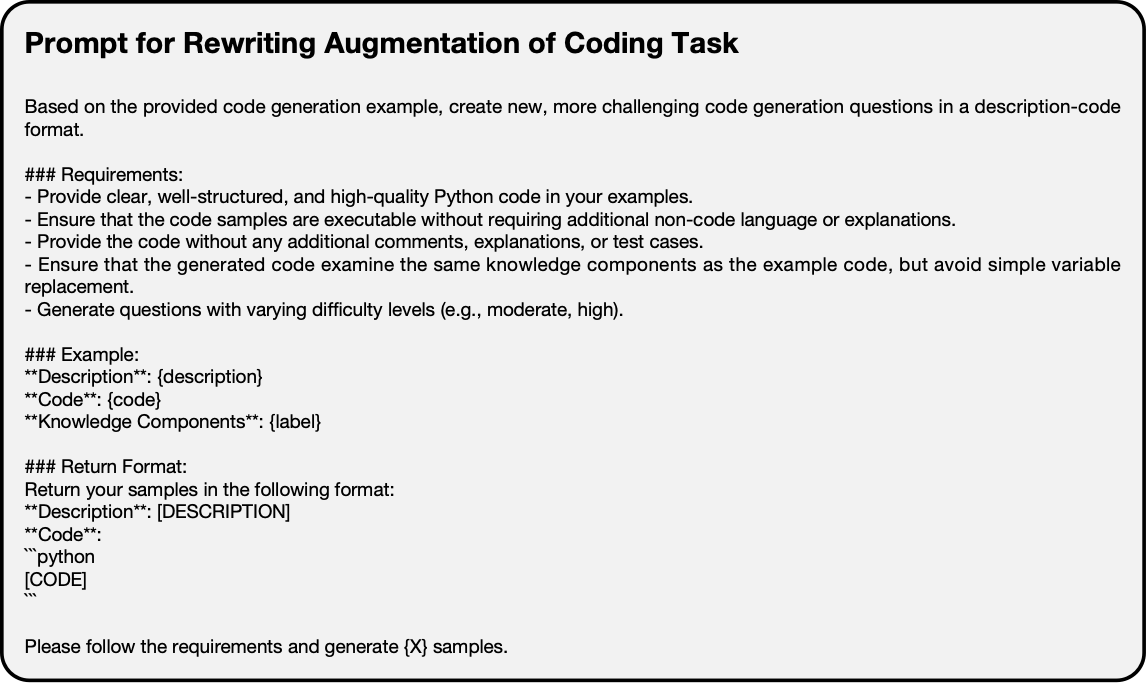}
  \caption{Prompt for Rewriting Augmentation of Coding Task.}
  \label{fig:14}
\end{figure*}

\begin{figure*}[htb]
  \centering
  \includegraphics[width=1.0\textwidth]{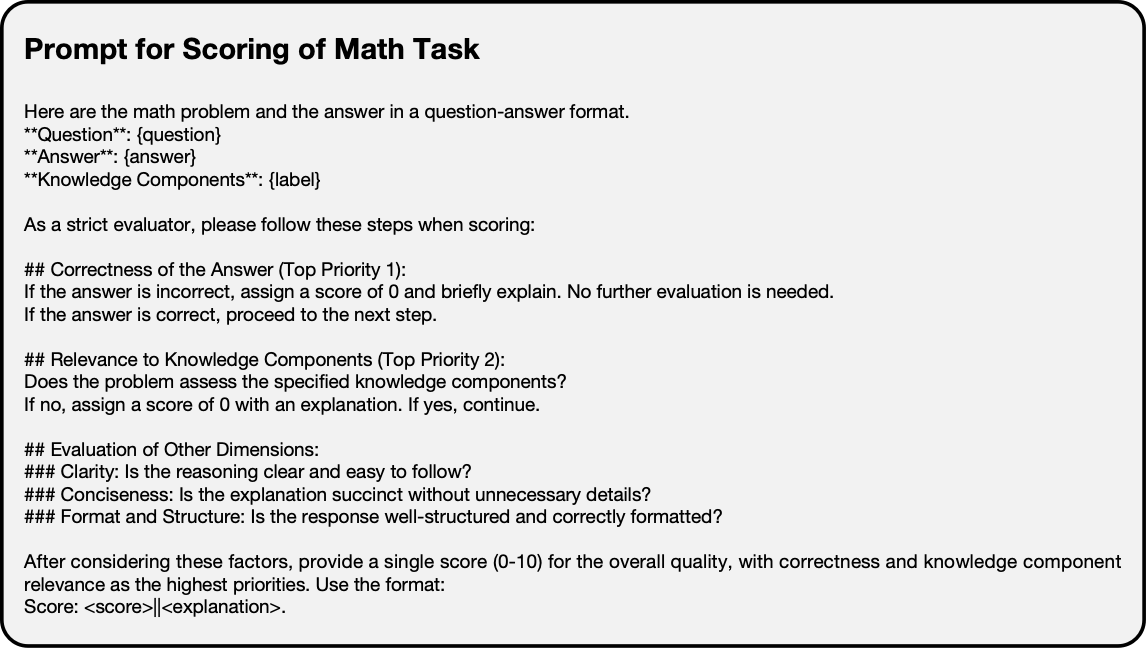}
  \caption{Prompt for Scoring of Math Task.}
  \label{fig:15}
\end{figure*}

\end{document}